\pdfoutput=1

\documentclass[11pt]{article}

\usepackage{naacl2021}
\usepackage{graphicx}
\usepackage{times}
\usepackage{latexsym}

\usepackage[T1]{fontenc}

\usepackage[utf8]{inputenc}

\usepackage{microtype}

\usepackage{booktabs}
\usepackage{multirow}
\usepackage{pifont}%
\usepackage{amsfonts}
\usepackage{bbm}

\title{Video Question Answering with Phrases via Semantic Roles}

\author{Arka Sadhu$^1$ \quad \quad Kan Chen$^{2}$ \quad \quad Ram Nevatia$^1$\\
$^1$University of Southern California \quad \quad $^2$Facebook Inc.\\
{\tt\small {\{asadhu|nevatia\}@usc.edu} \quad kanchen18@fb.com}
}

\begin{document}
\maketitle

\newcommand{\arka}[1]{}

\newcommand{\txi}[1]{{\texttt{#1}}}
\newcommand{\txbu}[1]{\underline{\textbf{#1}}}

\newcommand{\vdqa}[0]{VidQA}
\newcommand{\tk}[0]{VidQAP}
\newcommand{\tnamed}[1]{#1-QAP}
\newcommand{\tnamem}[1]{#1-QAP}

\newcommand{\anetdsn}[0]{ASRL-QA}
\newcommand{\chdsn}[0]{Charades-SRL-QA}

\newcommand{\cmark}{\ding{51}}%
\newcommand{\xmark}{\ding{55}}%

\newcommand{\best}[1]{\textbf{\underline{#1}}}
\newcommand{\sbest}[1]{\textbf{#1}}

\newcommand{\gh}[0]{\href{https://github.com/TheShadow29/Video-QAP}{https://github.com/TheShadow29/Video-QAP}}

\begin{abstract}
Video Question Answering (\vdqa{}) evaluation metrics have been limited to a single-word answer or selecting a phrase from a fixed set of phrases.
These metrics limit the \vdqa{} models' application scenario.
In this work, we leverage semantic roles derived from video descriptions to mask out certain phrases, to introduce \tk{} which poses \vdqa{} as a fill-in-the-phrase task.
To enable evaluation of answer phrases, we compute the relative improvement of the predicted answer compared to an empty string.
To reduce the influence of language-bias in \vdqa{} datasets, we retrieve a video having a different answer for the same question. 
To facilitate research, we construct ActivityNet-SRL-QA and Charades-SRL-QA and benchmark them by extending three vision-language models. 
We further perform extensive analysis and ablative studies to guide future work. 
\end{abstract}

\section{Introduction}
\label{sec:intro}

Given a video, Video Question Answering (\vdqa{}) requires a model to provide an answer to a video related question.
However, existing works treat \vdqa{} as an N-way ($N {\sim} 1k$) classification task across a fixed set of phrases. 
Models trained under such formulations are strictly restricted in their recall rate,  generalize poorly, and have severe limitations for end-user applications. 

\begin{figure}[!ht]
    \centering
    \includegraphics[width=\linewidth]{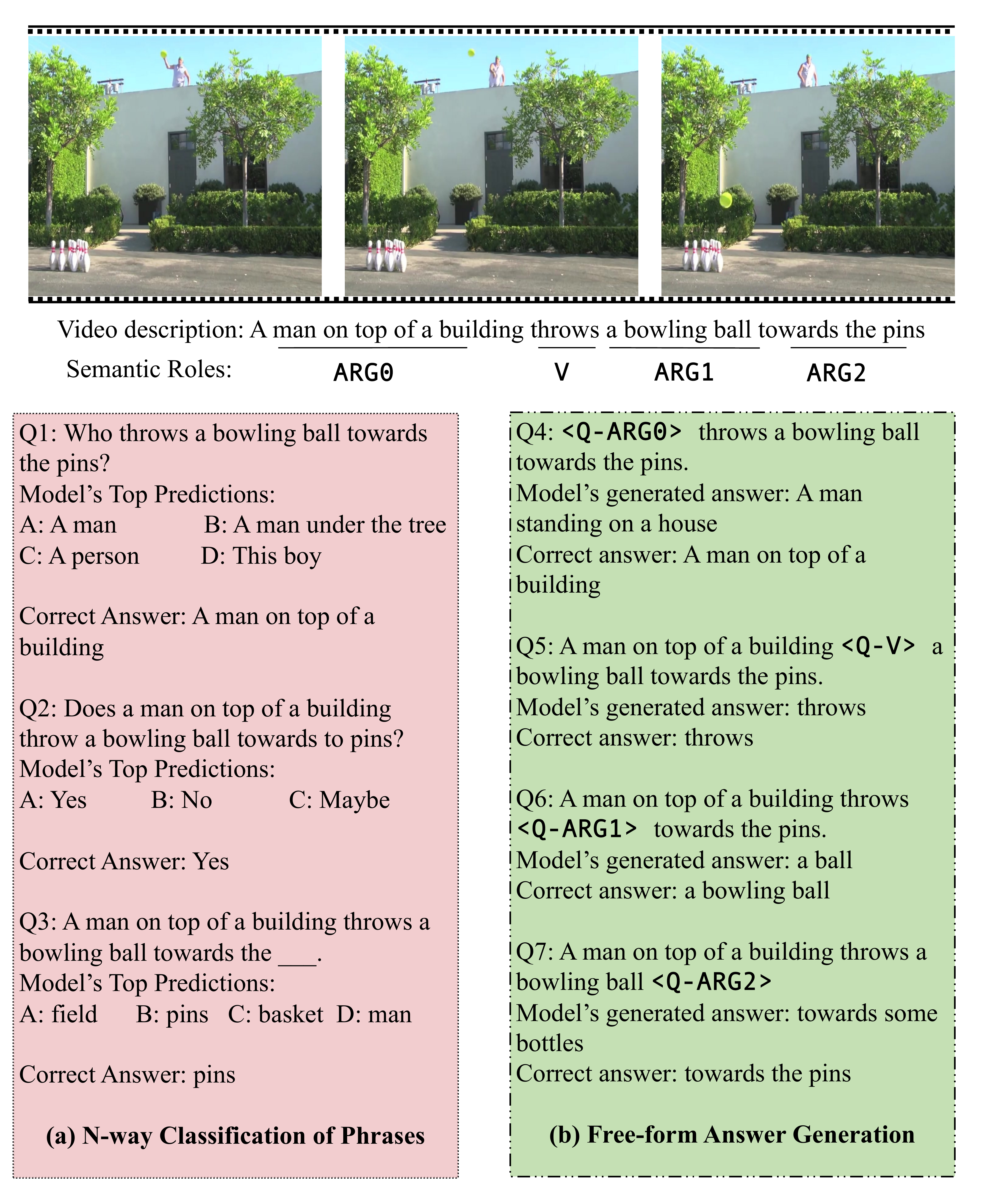}
    \caption{\footnotesize Previous methods formulate \vdqa{}  as a N-way classification task. The questions are converted via question generation tool (Q1, Q2) or masking-out strategy (Q3). However, such QA has a theoretical recall upper bound when the correct answer is not among the choice list. 
    In comparison, we propose a free-form text generation task which do not suffer such limitation (Q4-Q7)}
    \label{fig:intro_fig}
\end{figure}

In this work, we introduce Video Question Answering with Phrases (\tk{}) which treats \vdqa{} as a \textit{fill-in-the-phrase} task.
Instead of a question, the input to \tk{} consists of a query expression with a query-token.
Then, given a video, \tk{} requires replacing query-token with a sequence of generated words.
To generate a query, we leverage video descriptions and assign semantic roles to each phrase in these descriptions. 
Replacing a particular semantic-role with a query token produces a query-answer pair.
We illustrate this in Figure \ref{fig:intro_fig} (details in Section \ref{ss:ques_gen}).

\begin{table*}[t]
\centering
\resizebox{\linewidth}{!}{
\begin{tabular}{@{}cccccccccc@{}}
\toprule
Dataset         & Source        & \#Clips & Clip Duration(s) & \#QA-Pairs & \# QA / Clip & Task Type     & Scripts & Box    & QA Pair Creation \\ \midrule
Movie-QA      & Movies          & 6771    & 202.7 & 6462    & 0.95  & MC     & \cmark & \xmark & Human   \\
Movie-FIB      & Movies          & 128,085 & 4.8   & 348,998 & 2.72  & OE     & \xmark & \xmark & Automatic   \\
VideoQA*        & Internet videos & 18100   & 45    & 174,775 & 9.66  & OE     & \xmark & \xmark & Automatic   \\
MSVD-QA        & Internet videos & 1,970   & 9.7   & 50,505  & 25.64 & OE     &   \xmark     &    \xmark    &  Automatic   \\
MSR-VTT-QA     & Internet videos & 10,000  & 14.8  & 243,680 & 24.37 & OE     &    \xmark    &     \xmark   &  Automatic   \\
TGIF-QA        & Tumblr GIFs     & 62,846  & 3.1   & 139,414 & 2.22  & OE+MC  & \xmark & \xmark & Human+Automatic \\
TVQA           & TV Show         & 21,793  & 76    & 152,545 & 7     & MC     & \cmark & \xmark & Human   \\
TVQA+          & TV Show         & 4200    & 61.5  & 29,383  & 7     & MC     & \cmark & \cmark & Human   \\
ActivityNet-QA* & Internet videos & 5800    & 180   & 58000   & 10    & OE     & \xmark       &  \xmark      &  Human   \\
            \midrule
\anetdsn{}        & Internet videos & 35805   & 36.2  & 162091  & 5.54  & OE + Phrase & \xmark & \cmark & Automatic   \\
\chdsn{} & Crowd-Sourced & 9513    & 29.85             & 71735      & 7.54         & OE  + Phrase & \xmark  & \xmark & Automatic            \\ \bottomrule
\end{tabular}%
}
\caption{\footnotesize Comparison of Existing datasets for \vdqa{} with our proposed \anetdsn{} and \chdsn{}. Here, OE = Open-Ended, MC = Multiple Choice. ``Scripts'': if answering questions requires access to scripts or subtitles. ``Box'': if dataset provides bounding box annotations. *: Includes Yes/No questions}
\label{tab:vid_ds_comp}
\end{table*}

While free-form answer generation is highly desirable, 
evaluating them is non-trivial due to two main challenges.
First, existing language generation metrics like BLEU \cite{Papineni2002BleuAM} or BERTScore \cite{Zhang2020BERTScoreET} operate on sentences rather than phrases. 
When applied to short phrases, in the absence of context, even close matches like ``A person'' and ``The man'' would be falsely rejected due to no n-gram overlap or poor contextual embeddings.
Second, natural language questions often have strong language priors making it difficult to ascertain if the model retrieved information from the video.

To propose a reasonable evaluation metric,
we revisit our \textit{fill-in-the-phrase} formulation. 
Since we know where exactly the generated answer fits in the original query, we can create a complete sentence. 
With this key insight, we propose \textbf{relative scoring}: 
using the description as reference sentence, we compute the metrics once by replacing the query-token once with the predicted answer phrase and once with an empty-string. 
The model's performance is measured by the
relative improvement from the predicted answer compared to the empty string.
In particular, substituting the answer phrase in the query expression allows the computing the contextual embeddings required by BERTScore.

To mitigate the language-bias issue, we emulate the procedure proposed by \cite{Goyal2017MakingTV} where for a given question, another image (or video in our case) is retrieved which has a different answer for the same question.
To retrieve such a video, we use a contrastive sampling method  \cite{Sadhu2020VideoOG} over the dataset by comparing only the lemmatized nouns and verbs within the semantic roles (SRLs).
We then propose \textbf{contrastive scoring} to combine the scores of the two answer phrases obtained from the contrastive samples (details on evaluation in Section \ref{ss:evaling}).

To investigate \tk{}, we extend three vision-language models namely, Bottom-Up-Top-Down \cite{Anderson2018BottomUpAT}, VOGNet \cite{Sadhu2020VideoOG} and a Multi-Modal Transformer by replacing their classification heads with a Transformer \cite{Vaswani2017AttentionIA} based language decoder.
To facilitate research on \tk{} we construct two datasets ActivityNet-SRL-QA (\anetdsn{}) and \chdsn{} and provide a thorough analysis of extended models to serve as a benchmark for future research (details on model framework in Section \ref{ss:mdl_frm} and dataset creation in Section \ref{ss:expt_datasets}).

Our experiments validate the merits of moving away from N-way classification, and further show even among sequence generation models there exists a large disparity in performance across semantic-roles (i.e. queries for some roles can be answered very easily compared to other roles).
Moreover, certain roles hardly benefit from vision-language models suggesting room for improvement.
Finally, we investigate the effects of relative scoring and contrastive scoring for \tk{} with respect to BertScore.

Our contributions in this work are two-fold: 
(i) we introduce \tk{} and propose a systematic evaluation protocol to leverage state-of-art language generation metrics and reduce language bias
(ii) we provide extensive analysis and contribute a benchmark on two datasets evaluated using three vision-language models.
Our code and dataset are publicly available. \footnote{\gh{}}

\section{Related Works}
\label{sec:relwork}
\textbf{Question Answering in Images} has received extensive attention in part due to its end-user applicability.
Key to its success has been the availability of large-scale curated datasets like VQA v2.0 ~\cite{Goyal2017MakingTV} for visual question answering
and GQA ~\cite{Hudson2019GQAAN} for relational reasoning.
To address the strong language priors, the datasets are balanced by retrieving images which given the same question lead to a different answer.
However, these procedures cannot be extended for \vdqa{} since crowd-sourcing to retrieve videos is expensive and there exists no scene-graph annotations for videos.
In this work, we perform the retrieval using lemmatized nouns and verbs of the semantic roles labels obtained from video descriptions to balance the dataset.

\textbf{Question Answering in Videos:} has garnered less attention compared to ImageQA.
A major bottleneck is that there is no principled approach to curating a \vdqa{} dataset which reflects the diversity observed in ImageQA datasets.
For instance, naively crowd-sourcing video datasets leads to questions about color, number which is same as ImageQA datasets and doesn't reflect any spatial-temporal structure.
To address this issue, TGIF-QA ~\cite{Jang2017TGIFQATS} and ActivityNet-QA ~\cite{yu2019activityqa} use a question-template to enforce questions requiring spatio-temporal reasoning but forgo the question diversity. 
An orthogonal approach is to combine \vdqa{} with movie scripts ~\cite{Tapaswi2016MovieQAUS} or subtitles ~\cite{Lei2018TVQALC}. However, this severely restricts the domain of videos. 
Moreover, recent works have noted that language-only baselines often outperform vision-language baselines ~\cite{Jasani2019AreWA, Yang2020WhatGT, Zellers2019FromRT}.
A separate line of related research has focused on scene-aware dialogue \cite{alamri2019audio}. Instead of a single annotator providing both questions and answers, the annotation procedure follows a two-player game setup with one player asking a question and the other player answering with the roles switching after each turn. However, the evaluation method utilizes recall metrics which require the set of phrases to be known apriori. As a result, it doesn't strictly measure the performance of free-form generation but rather how well the ground-truth answer is ranked given a competing set of phrases which is analogous to multiple-choice questions.

\textbf{Automatic Question Generation:} Due to the above limitations, the dominant approach to create large-scale \vdqa{} dataset has been automatic question generation from existing video descriptions which can be easily crowd-sourced.
Our proposed formulation of using SRLs to generate query-expressions falls in this category.
Prior works include VideoQA ~\cite{Zeng2017LeveragingVD}, MSR-VTT-QA and MSVD-QA ~\cite{xu2017video} which use a rule based question generator ~\cite{Heilman2009QuestionGV} to convert descriptions to questions and Movie-Fill-in-the-Blanks ~\cite{Maharaj2017ADA} which mask outs at most one word which could be a noun, adjective or verb in a sentence.
In comparison, our method poses \tk{} as fill-in-blanks but with phrases, explicitly asks questions about actions, and the answer phrases are not constrained to a fixed set.
As a result of this increased space of phrases, methods on existing datasets cannot be directly applied to \tk{}.
To enable further research, we contribute two datasets \anetdsn{} and \chdsn{}.
In Table \ref{tab:vid_ds_comp} we compare these with existing \vdqa{} datasets.

\textbf{SRL in Vision:} has been explored in the context of human object interaction ~\cite{Gupta2015VisualSR}, situation recognition ~\cite{yatskar2016}, and multi-media extraction ~\cite{Li2020CrossmediaSC}.
Most related to ours is the usage of SRLs for grounding ~\cite{Silberer2018GroundingSR} in images and videos ~\cite{Sadhu2020VideoOG}.
Our work builds on ~\cite{Sadhu2020VideoOG} in using SRLs on video descriptions, however, our focus is not on grounding. Instead, we use SRLs primarily as a query generation tool and use the argument as a question directive.

\section{Design Considerations for \tk{}} 
\label{sec:method}

The \tk{} task is conceptually simple: given a video and a query expression with a query-token, a model should output an answer phrase that best replaces the query-token.
This leads to three main design considerations:
(i) How to generate a query-expression from existing resources (Section \ref{ss:ques_gen})
(ii) How to evaluate the answer phrases returned by a model (Section \ref{ss:evaling})
(iii) What modeling framework choices enable \tk{} (Section \ref{ss:mdl_frm}).

\subsection{Using SRLs to Generate Queries for \tk{}}
\label{ss:ques_gen}
We first briefly describe semantic-role labels (SRLs)\footnote{Detailed discussion is provided in supplementary. A demo is available here: https://demo.allennlp.org/semantic-role-labeling}. 
Then we detail how SRLs are used to create \tk{} queries.

\textbf{Query Generation Using SRLs:} 
Semantic Role Labels (SRLs) provide a high-level label to entities extracted from a sentence in the form of who (\txi{ARG0}), did what (\txi{V}) to whom (\txi{ARG1}) \cite{Strubell2018LinguisticallyInformedSF}.
Other roles such as to whom / using what (\txi{ARG2}) and where (\txi{LOC}) are also common.
As a pre-processing step, we assign SRLs to video descriptions using a state-of-art SRL labeler \cite{shi2019simple}.
A particular description could consist of multiple verbs, in which case, we consider each verb and its associated SRLs independently.
For a particular semantic-role, we substitute the corresponding phrase with a query token to generate the query expression.
The replaced phrase is the corresponding answer.
Using this method we are able to generate multiple queries from a single description.
An added merit of using SRLs is that query phrases are centered around ``verb-phrases'' which are highly relevant to the video content.

\begin{figure}[t]
    \centering
    \includegraphics[width=\linewidth]{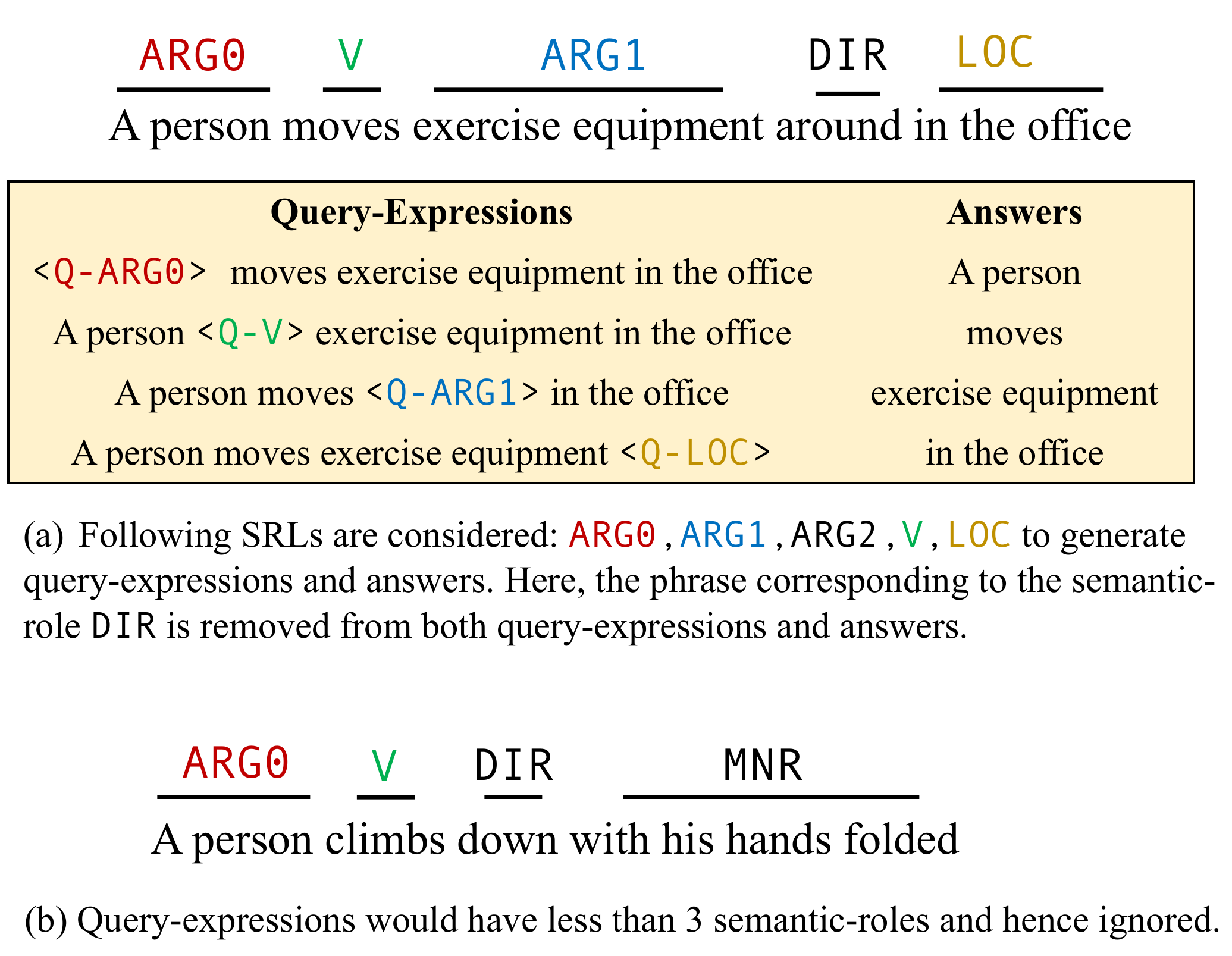}
    \caption{\footnotesize Illustration of our query generation process. In (a) \txi{DIR} is ignored from both Query and Answers. In (b) the question is removed from validation set since at most two arguments from considered set are present.} 
    \label{fig:qgen_fig}
\end{figure}

Generating queries using every SRL is not beneficial as some SRLs are more concerned with phrasing of the language rather than the video. 
For instance, in the phrase ``Players are running around on the field'', if we mask out the word ``around'' (\txi{DIR}), it can be answered without looking at the video.
To address the above issue, we confine our description phrases to a fixed set of semantic-roles namely: \txi{ARG0, ARG1, V, ARG2, ARGM-LOC}. 
Only those phrases which belong to the above set of SRLs may appear in the query-expression or as an answer phrase.
We further remove phrases which have only two arguments as these are too ambiguous to fill. 
Figure \ref{fig:qgen_fig} illustrates these steps. 

While using a slot for each slot could potentially limit the vocabulary used in each slot (for instance, the vocabulary set for ${<}\texttt{Q}{-}\texttt{ARG1}{>}$ could be limited to a small number of objects), empirically we don't find this to be the case (see Appendix \ref{ss:suppl_ds_stats} for detailed statistics). 
As a result, \tk{} is no simpler than \vdqa{} task.

We also remark that generating queries need not be strictly limited to masking out a single SRL and one could easily mask multiple SRLs in the same description.
However, we find two problems: first, for many cases, the output of masking multiple SRLs becomes exceedingly similar to video description task; second, using contrastive scoring (described in Section~\ref{ss:evaling}) for multiple SRLs becomes considerably more involved.
As a result, in this work, we focus on using a single SRL and keep the generalization to include multiple SRL queries for future work.

\subsection{Evaluating Answer Phrases}
\label{ss:evaling}
A key challenge in \tk{} is the lack of any standard protocol to evaluate free-form generated phrases.
A simple way is to adopt metrics like BLEU \cite{Papineni2002BleuAM}, ROUGE \cite{Lin2004ROUGEAP}, METEOR \cite{Banerjee2005METEORAA}, and CIDER \cite{Vedantam2015CIDErCI} which are already used for captioning in images and videos.
However, these metrics suffer from limited generalization: BLEU, ROUGE, and CIDER require exact n-gram matches. 
While this is fine for captioning where longer phrases average out errors, answers phrases are typically much smaller than a complete sentence. 
This leads to many near-correct answers receiving very low scores.
\begin{figure}[t]
    \centering
    \includegraphics[width=\linewidth]{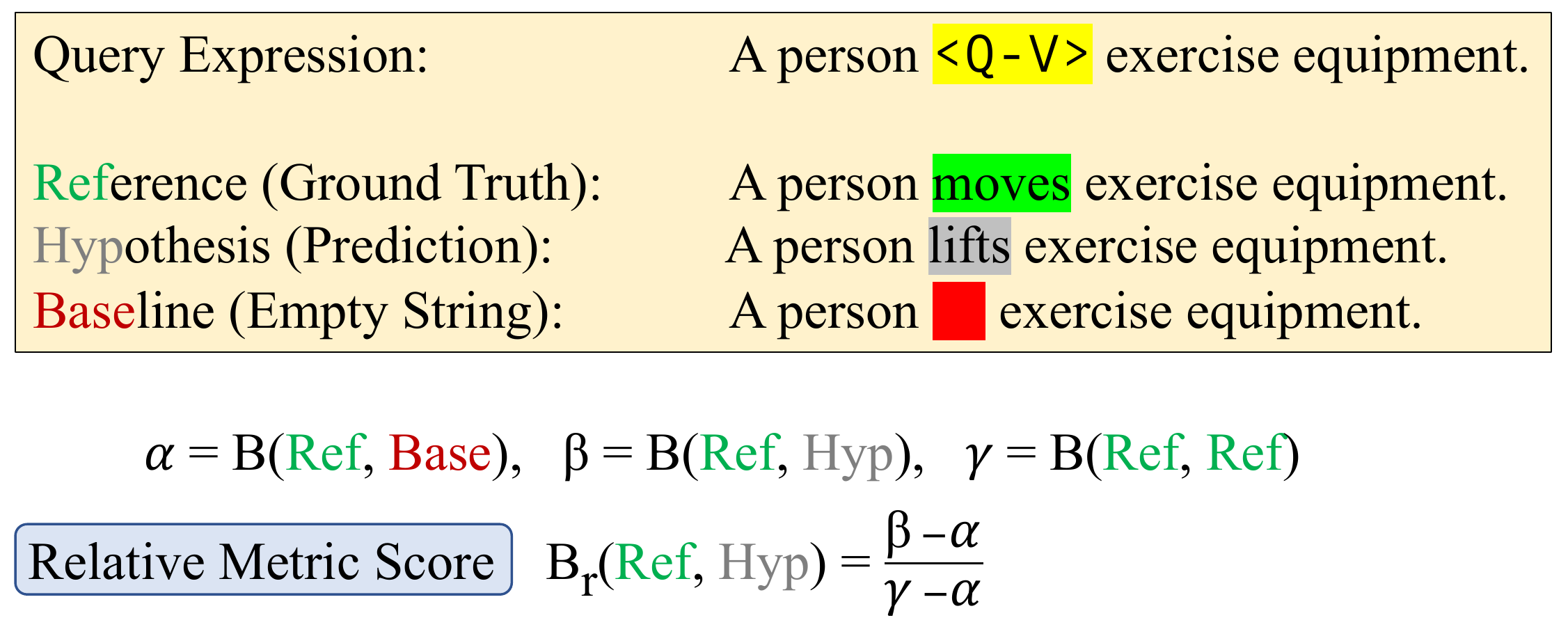}
    \caption{\footnotesize Illustration of the Relative Metric Computation. ``moves'' is the ground-truth answer and ``lifts'' is a model's prediction. 
    Relative Metric compares the relative improvement from using the model's prediction as compared to an empty string.} 
    \label{fig:evl_rescale}
\end{figure}

\begin{figure}[t]
    \centering
    \includegraphics[width=\linewidth]{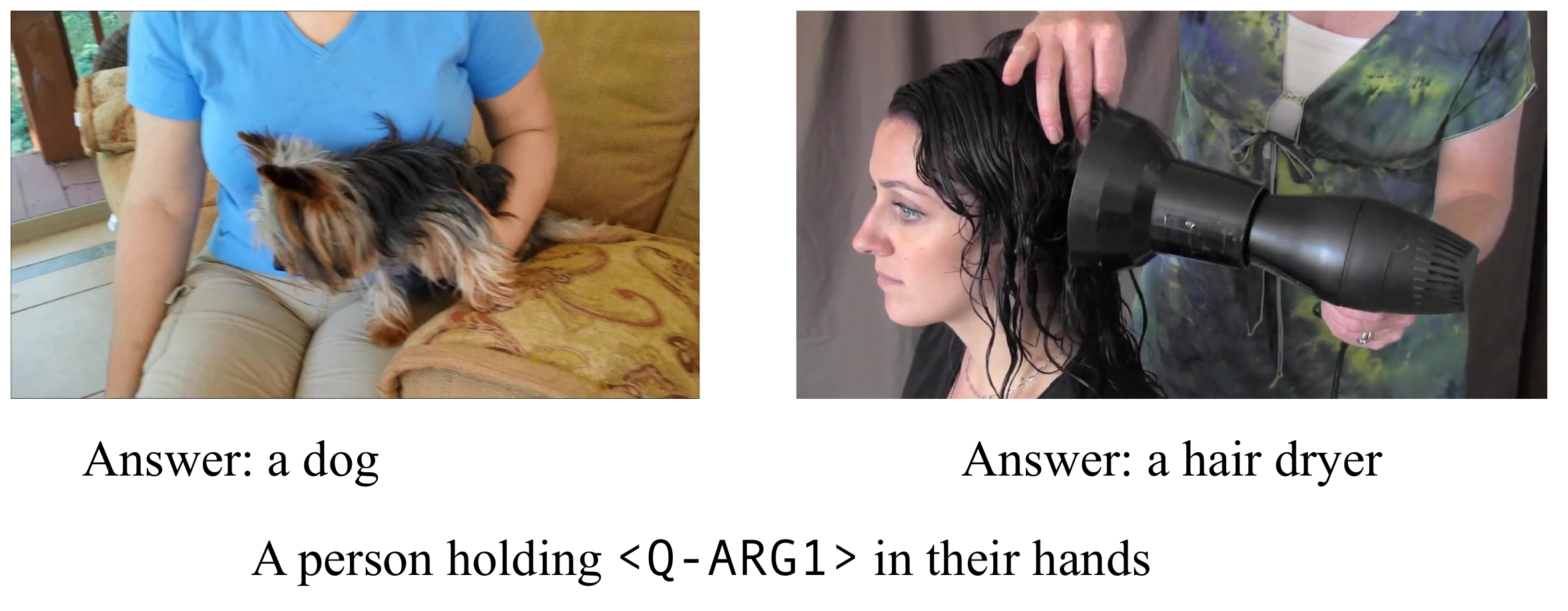}
    \caption{\footnotesize Illustration of Contrastive Sampling Process. For the same query-expression, we retrieve two videos with different answers. The model is required to correctly answer both the original and contrastive sample query.} 
    \label{fig:evl_cs}
\end{figure}

This issue is resolved to a certain extent for captioning by learned metrics like BERTScore \cite{Zhang2020BERTScoreET} which utilize contextual embeddings obtained from large pretrained models like BERT \cite{Devlin2019BERTPO} and RoBerta \cite{Liu2019RoBERTaAR}.
However, answer phrases are usually short and don't provide meaningful contextual embeddings. 
In the extreme case when the answer is a single word, for instance when the query is about a \txi{Verb}, these embeddings turn out to be very noisy leading to large number of false-positives.

\textbf{Relative Scoring:} 
To enable usage of contextual embeddings, we propose evaluating the relative improvement of the generated answer phrase compared to the ground-truth phrase.
We denote the input query expression as $Q$, the ground-truth answer is $A_{gt}$ ,and the predicted answer is $A_{pred}$.
Let $Q(X)$ denote $Q$ with the question tokens replaced by $X$. 
Then for a given metric $B$, we compute the relative metric $B_r$ as (see Figure \ref{fig:evl_rescale} for illustration)

    $Ref{=}Q(A_{gt}),
    Hyp{=}Q(A_{pred}),
    Base{=}Q(``")$

\begin{equation}
    \label{eqn:bs_rescale}
\resizebox{0.89\linewidth}{!}{
    $B_r(A_{gt}, A_{pred}) = \frac{B(Ref, Hyp) -  B(Ref, Base)}{B(Ref, Ref) - B(Ref, Base)}$
}
\end{equation}
Note that $B(Ref, Ref){=}1$ for BLEU, METEOR, ROUGE, BERTScore but not for CIDEr. 

The empty-string baseline in Eqn~\ref{eqn:bs_rescale} could be replaced with predictions from any model trained for this task.
In this work, we restrict to only empty-string baseline due to two desirable properties: its computational simplicity and it being agnostic to models and datasets. 

We further observe that Eqn \ref{eqn:bs_rescale} is very similar to the re-scaling proposed in BERTScore.
However, in BertScore re-scaling aims at making the score more readable and doesn't change the relative ranking of the hypothesis. 
In our case, Eqn \ref{eqn:bs_rescale} plays two roles:
first, it allows computing the contextual embeddings because the answers are now embedded inside a complete phrase, second while the ranking is not affected for a particular query, the score would be different across queries and hence affect the overall relative metric.

\begin{figure*}
    \centering
    \includegraphics[width=\linewidth]{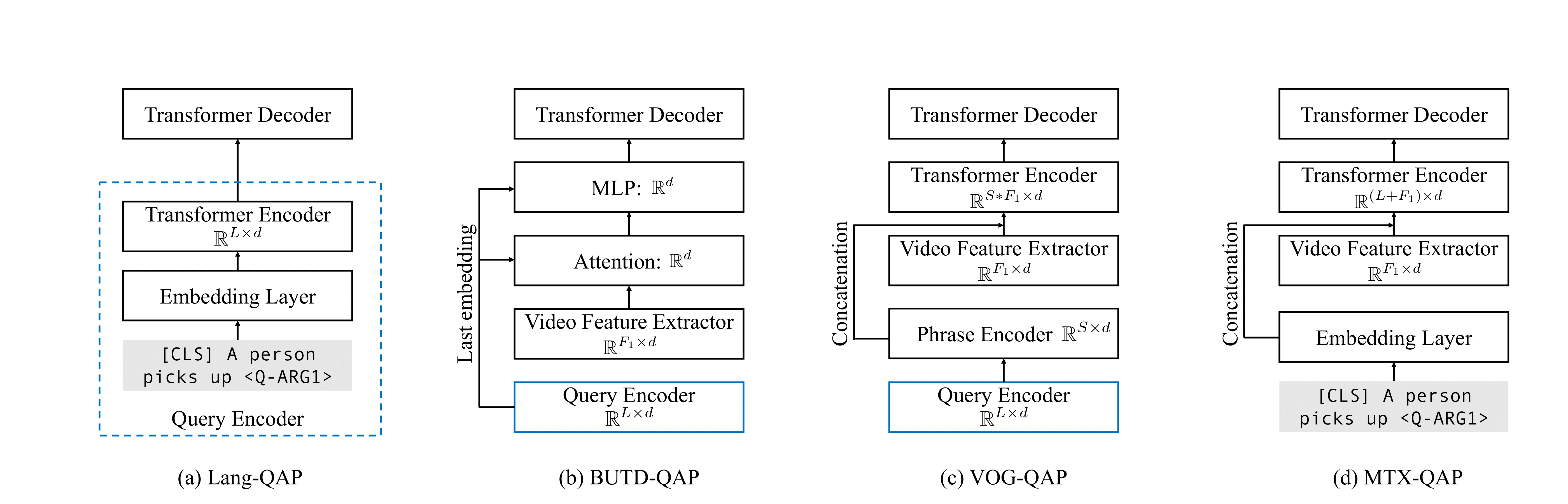}
    \caption{\footnotesize Schematic of the various models used to benchmark \tk{}. 
    Input Query: ``A person picks up \txi{<Q-ARG1>}''. Ground-Truth Answer: ``a pair of shoes''.
    (a) \tnamem{Lang} is a language-only model which encodes the query input and passes to a decoder. (b) \tnamem{BUTD} uses the pooled feature representation from language encoder and attends over the visual features. (c) \tnamem{VOG} uses an additional phrase encoder and applies a Transformer over the multi-modal features (d) \tnamem{MTX} consumes both the language and visual features with a multi-modal transformer.} 
    \label{fig:mdl_frm_fig}
\end{figure*}

\textbf{Contrastive Scoring:}
Visual Question Answering suffers from heavy language priors, and as a result, it is often difficult to attribute whether the image or video played a role in the success.
For images, \cite{Goyal2017MakingTV} resolved this by balancing the dataset where they crowd-sourced the task of collecting an image that has a different answer for the same question. 
However, such a crowd-sourcing method is difficult to extend to videos since searching for videos requires a much longer time. 
This is further complicated by accepting answer phrases compared to single word.

We simulate the balancing process using the contrastive sampling method used in \cite{Sadhu2020VideoOG}. 
Specifically, for a given video-query-answer $(V_1, Q_1, A_1)$ tuple we retrieve another video-query-answer $(V_2, Q_2, A_2)$ tuple which share the same semantic role structure as well as lemmatized noun and verbs for the question, but a different lemmatized noun for the answer.
At test time, the model evaluates the question separately, but the evaluation function requires both answers to be correct.
Since our answers comprise of phrases, the notion of correctness is not absolute (unlike say accuracy metric). 
Thus, we put a threshold $t$ below which the answer is deemed incorrect.

Mathematically, let $S_i {=} B_r(A_{gt_i}, A_{pred_i})$ be the relative score for sample $i$, and we are given sample $j$ is a contrastive example for sample $i$. 
Then the contrastive score ($CS_i$) for sample $i$ at a threshold $T_{CS}$ would be
\begin{equation}
    \label{eqn:cont_sampl_out}
    CS_i = max(S_i \mathbbm{1}[S_j > T_{CS} * B(Ref_j, Ref_j)], 0)
\end{equation}

Here $\mathbbm{1}[]$ is the indicator variable which is $1$ if the expression within brackets is True, otherwise $0$.
The $max$ operator ensures the scores don't become negative.
For our experiments, we use $T_{CS}{=}0$ which requires that the answer for the contrastive sample should be better than an empty string.

We further use the contrastive samples to compute a consistency metric.
For sample $i$, the consistency $Cons_i$ for a threshold $T_{cons}$ is given by 
\begin{equation}
    \label{eqn:cons_out}
    Cons_i = \mathbbm{1}[(S_i - T_{cons}) * (S_j - T_{cons}) > 0]
\end{equation}

As such, Consistency requires the model to be either correct or incorrect for both the original and the contrastive sample.

\textbf{Combined Metric at a Glance:}
Given metric $B$, for a given sample $i$ and contrastive sample $j$
\begin{enumerate}
    \itemsep0em
    \item Compute relative metric (Eqn \ref{eqn:bs_rescale}) for $i, j$
    \item Compute contrastive score (Eqn \ref{eqn:cont_sampl_out})
    \item Optionally compute Consistency (Eqn \ref{eqn:cons_out})
\end{enumerate}

We use the prefix ``R-'' such as in R-B to denote both relative scoring and contrastive scoring is being computed.
We report Consistency for BertScore with $T_{cons} {=} 0.1$

We note that, by construction, the relative scoring (Eqn~\ref{eqn:bs_rescale}) is positively correlated with human judgment, as the closer, the hypothesis is to the reference, the higher would the score be. 
The contrastive scoring is a metric used to prevent the model from guessing the correct answer by exploiting language biases and instead use the video to give a suitable prediction. Since humans don't have the ability to exploit such biases, it is difficult to relate to human evaluation.

\subsection{Model Framework}
\label{ss:mdl_frm}

Models for \tk{} require a language encoder to encode the question, a visual encoder to extract video features, a multi-modal module to jointly learn over vision-language space and a decoder to generate a sequence of words.

\textbf{Inputs} include query expression $\{w\}_{i=1}^L$ ($L$ is number of words), video segment features for $F_1$ frames and optionally $k$ RCNN features for $F_2$ frames. 
In either case, frames are sampled uniformly from the video segment time-span.
While the models differ in their encoding scheme, our language decoder model (Transformer based) used to generate the output answer phrase is kept same across all models with QAP suffix.

\txbu{\tnamem{Lang}}: is a language-only (video-blind) model using only the query input. 
It uses Transformer based encoder to encode the query into $\hat{q} \in \mathbb{R}^{L \times d}$. 
The decoder subsequently uses the last layer output of the encoder (Figure\ref{fig:mdl_frm_fig}-(a)).

\txbu{\tnamem{BUTD}}: Bottom-up-Top-Down \cite{Anderson2018BottomUpAT} is a popular approach for image question answering as well as captioning. 
It first computes attention between the question and the RCNN visual features to generate an attended visual feature, which is then used with the question to produce an output answer.
Here, we replace the RCNN features with the segment features ($\hat{v} \in \mathbb{R}^{F_1 \times d}$). 
We can also include RCNN features by projecting them to same dimension as segment features and then concatenate them along the frame-axis ($\hat{v} \in \mathbb{R}^{(F_1+F_2*k) \times d}$).
For language features, we use the [CLS] token representation from the last layer of the language encoder used in \tnamem{Lang}. 
The output using the language and visual features is ($\hat{m} \in \mathbb{R}^{d}$) passed to the decoder (Figure \ref{fig:mdl_frm_fig}(b)).

\txbu{\tnamem{VOG}}: VOGNet \cite{Sadhu2020VideoOG} has been proposed for grounding objects in videos given a natural language query. 
Following the architecture, we first derive phrase encoding which corresponds to a single SRL i.e. $\hat{q} \in \mathbb{R}^{S \times d}$ ($S$ is number of semantic roles).
These phrase features are concatenated with the visual features (same as those used in \tnamem{BUTD} (i.e. $\hat{v}$)) to get multi-modal features $m[l, i] {=} [\hat{v}_i || \hat{q}_l]$ and then reshaped to get $m \in \mathbb{R}^{S * F \times d}$. 
These multi-modal features are subsequently passed to decoder to generate the output sequence (Figure \ref{fig:mdl_frm_fig} (c)).

\txbu{\tnamem{MTX}}: 
Recently, transformer models pre-trained on large-scale paired image-text data have become popular.
Even in the absence of pre-training, such architectures can achieve competitive performance \cite{Lu2019ViLBERTPT}.
In the context of videos, ActBert \cite{Zhu2020ActBERTLG} has been proposed.
We create a similar architecture to ActBert but we replace their proposed Tangled-Transformer with a vanilla Transformer \footnote{The code for ActBert is not publicly available.}.
Specifically, we jointly encode the language and visual features in a single transformer and feed the output to the decoder (Figure \ref{fig:mdl_frm_fig} (d)).

\txbu{LangCL and MTxCL:}
Apart from QAP models, we also consider their phrase classification counterparts where the decoder is replaced with a N-way classifier (two-layered MLP in our case) across a fixed set of phrases.
For our experiments, we used $N{=}1k$ phrases for LangCL and $N{\in}\{1k, 10k\}$ for MTxCL.

\section{Experiments}
\label{sec:expts}

We briefly discuss the dataset creation process (Section \ref{ss:expt_datasets}), followed by experimental setup (Section \ref{ss:expt_setup}).
We then summarize our results (Section \ref{ss:expt_results}) and discuss key-findings.
We provide implementation details, qualitative visualizations of our dataset, metrics and trained models in the appendix.

\subsection{Dataset Creation}
\label{ss:expt_datasets}

We create two datasets \anetdsn{} and \chdsn{} derived from ActivityNet-Captions \cite{Krishna2017DenseCaptioningEI} and Charades \cite{Sigurdsson2016HollywoodIH} respectively.

There are three key steps to create QA datasets from descriptions: 
(i) assign semantic-roles to the descriptions 
(ii) perform co-reference resolution so that the questions are self-contained 
(iii) obtain lemmatized nouns and verbs to perform contrastive sampling.
For semantic-role labeling, we use \cite{shi2019simple}.
For co-reference resolution, we use the co-reference resolution model provided by allennlp library \cite{Gardner2017AllenNLP} which uses the model by \cite{Lee2017EndtoendNC} but replaces the GloVe \cite{Pennington2014GloveGV} embeddings with SpanBERT embeddings \cite{Joshi2019SpanBERTIP} \footnote{https://demo.allennlp.org/coreference-resolution}.

Since Charades primarily involves videos with a single person, we discard questions involving \txi{ARG0}.
We limit to using a single description per video to avoid repetitive questions.
We re-use the same train split for both datasets.
For ASRL-QA, test set of ActivityNet is not public and Charades only has a test set but no official validation set.
Thus, we split the existing validation set by video names and create the validation and test sets.
For both validation and test splits, we remove those questions for which no contrastive sample was found as it indicates data-biases.

\subsection{Experimental Setup}
\label{ss:expt_setup}

\textbf{Dataset Statistics:} 
\anetdsn{} has 35.7k videos and 162k queries split into train, validation and test sets with 30.3k, 2.7k, 2.7k videos and 147k, 7.5k, 7.5k queries. 
We observe that the size of validation and test sets are proportionately smaller compared to their respective train sets. 
This is because only queries with corresponding contrastive sample are included while no such filtering is done for the train set (${\sim}95k$ queries in train set have a contrastive pair).  
\chdsn{} contains 9.4k videos and 71.7k queries split across train, validation and test sets with 7.7k, 0.8k, 0.8k videos and 59.3k, 6.1k, 6.2k queries.
Despite its smaller size, the size of validation, test sets of \chdsn{} is comparable to \anetdsn{} as Charades is curated with the goal of diversifying subject, verb, object tuples. 
Supplementary material provides further details on 
the dataset statistics and visualizations.

\textbf{Evaluation Metrics:} 
As discussed in Section \ref{ss:evaling}, we report the combined metric (i.e. metrics prefixed with ``R-'') for the commonly used generation metrics: BLEU, METEOR, ROUGE, CIDEr and BertScore (implementations from ~\cite{Chen2015MicrosoftCC,Zhang2020BERTScoreET}).
For BLEU, we report the sentence level BLEU-2.
All reported results are test set results using the model which performs best on validation set.

\begin{table*}[t]
\centering
\resizebox{0.88\linewidth}{!}{
\begin{tabular}{@{}c|cccccc|cccccc@{}}
\toprule
         & \multicolumn{6}{|c|}{\anetdsn}           & \multicolumn{6}{c}{\chdsn}   \\ 
         & R-BS & Cons & R-B@2 & R-R & R-M & R-C & R-BS & Cons & R-B@2 & R-R & R-M & R-C \\ \midrule
LangCL ($1k$)        & 0.253 & \best{0.889} & 0.120 & 0.098 & 0.071 & 0.044 & 0.293 & 0.697 & 0.224 & 0.209 & 0.114 & 0.077 \\
MTxCL  ($1k$)        & 0.255 & \sbest{0.869} & 0.130 & 0.114 & 0.080 & 0.050 & 0.288 & 0.707 & 0.215 & 0.208 & 0.116 & 0.075 \\ 
MTxCL  ($10k$) & 0.286 & 0.788 & 0.157 & 0.133 & 0.100 & 0.061 & 0.408 & 0.695 & 0.286 & 0.261 & 0.142 & 0.108 \\
\midrule
\tnamem{Lang} & 0.402 & 0.728 & 0.228 & 0.182 & 0.125 & 0.095 & 0.406 & 0.719 & 0.277 & 0.253 & 0.147 & 0.121 \\
\tnamem{BUTD} & 0.413 & 0.716 & 0.237 & 0.203 & \sbest{0.147} & 0.105 & 0.399 & 0.714 & 0.271 & 0.231 & 0.115 & 0.105 \\
\tnamem{VOG}  & \best{0.414} & 0.717 & \sbest{0.239} & \sbest{0.204} & 0.142 & \sbest{0.108} & \best{0.442} & \sbest{0.739} & \best{0.297} & \best{0.274} & \best{0.165} & \sbest{0.136} \\
\tnamem{MTX}  & \sbest{0.414} & 0.715 & \best{0.247} & \best{0.206} & \best{0.149} & \best{0.113} & \sbest{0.439} & \best{0.757} & \sbest{0.294} & \sbest{0.267} & \sbest{0.157} & \best{0.139} \\ \bottomrule
\end{tabular}
}
\caption{\footnotesize Comparison of our extended models for \tk{} and Classification based (CL) models across two datasets on our proposed Metric. Here, ``R-" prefix implies it is the final metric computed after relative scoring and contrastive scoring with threshold $0$. ``BS": BertScore, ``Cons'': Consistency on BertScore, B@2: Sentence BLEU-2, R: ROUGE, M: METEOR, C: CIDEr. Reported numbers are on the test set.
For classification models, the number within the parenthesis denotes the size of fixed vocabulary of phrases.
\best{Best result},  \sbest{Second Best result}.
}
\label{tab:phrase_scores_main}
\end{table*}

\begin{table}[t]
\centering
\resizebox{\linewidth}{!}{
\begin{tabular}{@{}c|ccccc|cccc@{}}
\toprule
         & \multicolumn{5}{c|}{\anetdsn}       & \multicolumn{4}{c}{\chdsn} \\ 
         & ARG0 & V & ARG1 & ARG2 & LOC & V    & ARG1   & ARG2   & LOC   \\
         \midrule 
LangCL ($1k$)        & 0.598 & 0.423 & 0.102 & 0.125 & 0.018 & 0.564 & 0.291 & 0.146 & 0.173 \\
MTxCL  ($1k$)        & 0.607 & 0.399 & 0.106 & 0.142 & 0.019 & 0.549 & 0.346 & 0.152 & 0.106 \\
MTxCL  ($10k$) & 0.697 & 0.379 & 0.161 & 0.144 & 0.049 & 0.601 & 0.445 & 0.315 & 0.272 \\
\midrule

\tnamem{Lang} & \sbest{0.697} & \best{0.519} & 0.325 & 0.322 & 0.145 & \sbest{0.631} & 0.458 & 0.33  & 0.206 \\
\tnamem{BUTD} & 0.681 & \sbest{0.515} & \sbest{0.372} & \sbest{0.334} & 0.162 & 0.568 & 0.413 & 0.316 & 0.299 \\
\tnamem{VOG}  & 0.671 & 0.513 & 0.366 & 0.332 & \best{0.188} & 0.63  & \best{0.467} & \best{0.365} & \best{0.305} \\
\tnamem{MTX}  & \best{0.702} & 0.478 & \best{0.374} & \best{0.344} & \sbest{0.17}  & \best{0.633} & \sbest{0.455} & \sbest{0.364} & \sbest{0.304} \\ \bottomrule
\end{tabular}%
}
\caption{\footnotesize Comparison of our extended models per SRL. All reported scores are R-BS: BertScore computed after relative scoring and contrastive scoring with threshold 0.}
\label{tab:per_arg_bscore}
\end{table}

\subsection{Results and Discussions}
\label{ss:expt_results}

Table \ref{tab:phrase_scores_main} compares performance of the proposed \tk{} models with N-way classification baselines (denoted with suffix ``CL'') on \anetdsn{} and \chdsn{}.

\textbf{Comparing Metrics:} 
It is evident that compared to other metrics, R-BertScore shows a higher relative improvement.
This is because BertScore allows soft-matches by utilizing contextual embeddings obtained from a pre-trained BERT \cite{Devlin2019BERTPO} or Roberta \cite{Liu2019RoBERTaAR} model.

\textbf{Comparison Across Datasets:} 
We find that performance on both datasets follow very similar trends across all metrics.
\chdsn{} has slightly higher scores compared to \anetdsn{} likely because it has lesser data variations (Charades is mostly confined indoor videos) suggesting findings on either dataset would transfer.

\begin{table}[t]
\centering
\resizebox{0.92\linewidth}{!}{
\begin{tabular}{@{}ccccccc@{}}
\toprule
      &            & ARG0  & V      & ARG1  & ARG2  & LOC \\ \midrule
\multirow{6}{*}{\rotatebox[origin=c]{90}{\tnamem{Lang}}}  
 & Direct   & 0.552 & 0.9268 & 0.234 & 0.302 & 0.216    \\
 & Rel Score & 0.7   & 0.534  & 0.332 & 0.237 & 0.1      \\
 & CS@0    & 0.697 & 0.519  & 0.325 & 0.322 & 0.145    \\
 & CS@0.1  & 0.69  & 0.492  & 0.295 & 0.28  & 0.132    \\
 & CS@0.2  & 0.68  & 0.459  & 0.262 & 0.212 & 0.106    \\
 & CS@0.3  & 0.657 & 0.423  & 0.219 & 0.149 & 0.085    \\
      \midrule
      \midrule
\multirow{6}{*}{\rotatebox[origin=c]{90}{\tnamem{MTX}}} 
 & Direct   & 0.566 & 0.929 & 0.269 & 0.321 & 0.258    \\
 & Rel Score & 0.706 & 0.488 & 0.366 & 0.25  & 0.14     \\
 & CS@0    & 0.702 & 0.478 & 0.374 & 0.344 & 0.17     \\
 & CS@0.1  & 0.693 & 0.45  & 0.343 & 0.305 & 0.145    \\
 & CS@0.2  & 0.681 & 0.413 & 0.306 & 0.239 & 0.117    \\
 & CS@0.3  & 0.659 & 0.376 & 0.27  & 0.17  & 0.08     \\ \bottomrule

\end{tabular}%
}
\caption{\footnotesize BertScore Metrics computed Directly on answer phrases. Rel Score: After Relative Scoring. CS@T: Contrastive scoring with threshold T. }
\label{tab:evl_rescale_cs}
\end{table}
\begin{table}[t]
\centering
\resizebox{\linewidth}{!}{
\begin{tabular}{@{}ccccccc@{}}
\toprule
     & ARG0  & V     & ARG1  & ARG2  & LOC   & Overall \\ \midrule
\tnamem{BUTD} & \best{0.706} & \sbest{0.506} & \best{0.388} & \best{0.36}  & \sbest{0.196} & \best{0.431}   \\
\tnamem{VOG}  & \sbest{0.704} & \best{0.516} & 0.366 & 0.352 & \best{0.202} & \sbest{0.429}   \\
\tnamem{MTX}  & 0.685 & 0.465 & \sbest{0.378} & \sbest{0.355} & 0.19  & 0.416   \\ \bottomrule
\end{tabular}%
}
\caption{\footnotesize Effect of Adding Region Proposals. All reported scores are R-BS. 
\best{Best result},  \sbest{Second Best result}.
}
\label{tab:add_props}
\end{table}

\textbf{Comparison within N-way Classification:}
We notice that when $1k$ fixed set of phrases are used classification models show very limited performance.
Allowing $10k$ phrases gives a significant improvement in performance on \chdsn{} ($12$ points on R-BS) however this doesn't translate to \anetdsn{}.
This is because \anetdsn{} contains many more probable phrases ($29K$ compared to $8K$) in their respective training sets.
We also notice that increasing the number of phrases vocabulary coincides with decreasing consistency.

\textbf{Comparing Free-from Answer Generation (QAP) with N-way Classification (CL):}
We investigate the advantages of using a decoder network to generate phrases compared to an N-way classification over a fixed set of phrases (denoted with the suffix ``CL'' and number of phrases used in parenthesis).
Table ~\ref{tab:phrase_scores_main} shows that both Lang-QAP and MTX-QAP outperform their classification counterparts, namely Lang-CL and MTX-CL on both datasets. 
This implies the free-form generation are not limited to simply generating the most frequently appearing phrases in the training set, thereby showing its effectiveness.

\textbf{Comparison Across Models:}
We find that multi-modal models outperform language-only baseline. 
However, the improvement over language baseline is small.
The reason for the small gap is elucidate in Table \ref{tab:per_arg_bscore} where we report R-BertScore for every considered SRL.

We find a large disparity in performance depending on the SRL. 
Most strikingly, multi-modal models perform worse than language-only model on \txi{ARG0} and \txi{V}.
For \txi{ARG0}, the strong performance of the \tnamem{Lang} arises because most of the time the agent who causes an action is a human. 
Therefore answer phrases having simply ``A man'' or ``A woman'' or ``A person'' leads to reasonable performance.
This additionally suggests that grounding ``who'' is performing the action remains non-trivial.

The more surprising result is the strong performance of \tnamem{Lang} on \txi{V} which is consistent across both datasets despite using contrastive sampling. 
There are two likely causes. 
First, the distinction between verbs is not as strict as object nouns, i.e. even similar verbs are classified as a separate verb diminishing the returns of contrastive sampling. 
For instance, ``jumping'' and ``hoping'' have different lemma and thus considered distinct verbs but R-BS would treat them as similar even if the specific action would be classified ``jumping'' rather than ''hoping''.
Second, SRLs such as \txi{ARG1} confines the set of possible verbs.
For instance, if the object is ``glass'', only limited verbs such as ``drink'', ``hold'' are probable.

On the remaining arguments namely \txi{ARG1}, \txi{ARG2}, and \txi{LOC}, multi-modal models show a steady improvement over language-only baseline ranging from $1{-}10\%$.
However, the performance in absolute terms remains very low.
As such, our proposed task \tk{} remains extremely challenging for current multi-modal models.

\textbf{Evaluation Metric Scores:}
In Table \ref{tab:evl_rescale_cs} we record the BertScore computation in three parts: 
directly computing over the answer phrases, 
performing relative scoring, 
finally performing contrastive scoring with different thresholds.

We observe that for \txi{V}, naive computation leads to absurdly high scores.
This is because verbs consist of a single word which means the embeddings are not contextual.
This is remedied by relative scoring and is further controlled by combining with contrastive sampling.

Further note that relative scoring operates differently based on the SRLs.
For instance, it increases the score for \txi{ARG0} and \txi{ARG1} where the answers  more often paraphrased the ground-truth questions while for \txi{ARG2} and \txi{LOC}, it decreases the score due to incorrect matches.
While contrastive scoring is aimed at reducing language-only bias and as such should always reduce the relative score, we observe increased score in \txi{ARG2} for both \tnamem{Lang} and \tnamem{MTX}. 
This is caused by the $max$ function which restricts the lower-limit to be $0$.

\textbf{Effect of Region Boxes:}
As noted earlier, the visual features can also include region features extracted from an object detector like FasterRCNN \cite{Ren2015FasterRT}. 
In Table \ref{tab:add_props} we record the effect of including regional features. 
In particular, we use the GT5 setting used in \cite{Sadhu2020VideoOG} where $5$ region proposals are used from $10$ frames uniformly sampled from the video segment. 
Interestingly, \tnamem{MTX} under-performs than both \tnamem{BUTD} and \tnamem{VOG} on \txi{ARG0}. 
A possible reason is that the transformer is unable to effectively reason over both language and vision over such a large range of inputs.

\section{Conclusion}
In this work, we introduce Video Question Answering with Phrases (\tk{}) where we pose \vdqa{} as a fill-in-the-phrase task.
Given a video and query expression, a model needs to compose a sequence of words to answer.
We then propose a method to leverage semantic roles from video descriptions to generate query expressions and outline a robust evaluation protocol.
This involves computing the relative improvement of the prediction answer compared to an empty string followed by a contrastive sampling stage which reduces language-only biases.
We then contribute two datasets \anetdsn{} and \chdsn{} to facilitate further on \tk{} and benchmark them with three vision-language models extended for our proposed task.
\section*{Acknowledgement}
We thank the anonymous reviewers for their suggestions and feedback. 
This research was supported, in part, by the Office of Naval Research under grant \#N00014-18-1-2050.
\section*{Ethics Statement}
In this work, we propose an extension to the existing video question answering framework to include free-form answers and suggest how to evaluate such a task.

\textbf{Direct Application (Positive):} 
A direct application of our task would be to enrich existing descriptions obtained from video captioning models which could lead to better video retrieval results. 
For instance, one could query about what tool to use in order to cut a piece of cardboard by querying ``A person cutting a piece of cardboard $<$Q-ARG2$>$".

\textbf{Direct Application (Negative):} 
Caution must be taken in directly applying models trained on descriptions without properly balancing the data-distributions as it is possible that hidden data-biases are amplified.
As an example, ASRL-QA has many videos involving men throwing shot puts. 
As a result, a model could learn this biased correlation and whenever queried ``who'' ($<$Q-ARG0$>$ throws a shot put) it would always produce the answer ``man'' even if the video clearly shows a ``woman''.

\textbf{Broader Societal Impacts (Positive):}
Question answering is an excellent tool for diagnosing a model's understanding due to its high interactivity. 
Our proposed formulation takes this a step forward with answer phrases and can in-turn facilitate human-computer interactions. 
Our proposed model can be extended to down-stream tasks such as retrieving a video or retrieving a part of the video given a question or query. 

\textbf{Broader Societal Impacts (Negative):}
Since our method is agnostic to the end user case, it can be re-purposed to extract out sensitive information and be a threat to privacy.
\newpage

\bibliography{ref}

\begin{thebibliography}{52}
\expandafter\ifx\csname natexlab\endcsname\relax\def\natexlab#1{#1}\fi

\bibitem[{Alamri et~al.(2019)Alamri, Cartillier, Das, Wang, Cherian, Essa,
  Batra, Marks, Hori, Anderson et~al.}]{alamri2019audio}
Huda Alamri, Vincent Cartillier, Abhishek Das, Jue Wang, Anoop Cherian, Irfan
  Essa, Dhruv Batra, Tim~K Marks, Chiori Hori, Peter Anderson, et~al. 2019.
\newblock Audio visual scene-aware dialog.
\newblock In \emph{Proceedings of the IEEE/CVF Conference on Computer Vision
  and Pattern Recognition}, pages 7558--7567.

\bibitem[{Anderson et~al.(2018)Anderson, He, Buehler, Teney, Johnson, Gould,
  and Zhang}]{Anderson2018BottomUpAT}
Peter Anderson, X.~He, C.~Buehler, Damien Teney, Mark Johnson, Stephen Gould,
  and Lei Zhang. 2018.
\newblock Bottom-up and top-down attention for image captioning and visual
  question answering.
\newblock \emph{2018 IEEE/CVF Conference on Computer Vision and Pattern
  Recognition}, pages 6077--6086.

\bibitem[{Baker et~al.(1998)Baker, Fillmore, and Lowe}]{baker1998framenet}
Collin~F Baker, Charles~J Fillmore, and John~B Lowe. 1998.
\newblock The berkeley framenet project.
\newblock In \emph{36th Annual Meeting of the Association for Computational
  Linguistics and 17th International Conference on Computational Linguistics,
  Volume 1}, pages 86--90.

\bibitem[{Banerjee and Lavie(2005)}]{Banerjee2005METEORAA}
S.~Banerjee and A.~Lavie. 2005.
\newblock Meteor: An automatic metric for mt evaluation with improved
  correlation with human judgments.
\newblock In \emph{IEEvaluation@ACL}.

\bibitem[{Bonial et~al.(2012)Bonial, Hwang, Bonn, Conger, Babko-Malaya, and
  Palmer}]{bonial2012propbank}
Claire Bonial, Jena Hwang, Julia Bonn, Kathryn Conger, Olga Babko-Malaya, and
  Martha Palmer. 2012.
\newblock English propbank annotation guidelines.
\newblock \emph{Center for Computational Language and Education Research
  Institute of Cognitive Science University of Colorado at Boulder}, 48.

\bibitem[{Chen et~al.(2015)Chen, Fang, Lin, Vedantam, Gupta, Doll{\'a}r, and
  Zitnick}]{Chen2015MicrosoftCC}
Xinlei Chen, H.~Fang, Tsung-Yi Lin, Ramakrishna Vedantam, Saurabh Gupta,
  P.~Doll{\'a}r, and C.~L. Zitnick. 2015.
\newblock Microsoft coco captions: Data collection and evaluation server.
\newblock \emph{ArXiv}, abs/1504.00325.

\bibitem[{Denkowski and Lavie(2014)}]{denkowski:lavie:meteor-wmt:2014}
Michael Denkowski and Alon Lavie. 2014.
\newblock Meteor universal: Language specific translation evaluation for any
  target language.
\newblock In \emph{Proceedings of the EACL 2014 Workshop on Statistical Machine
  Translation}.

\bibitem[{Devlin et~al.(2019)Devlin, Chang, Lee, and
  Toutanova}]{Devlin2019BERTPO}
J.~Devlin, Ming-Wei Chang, Kenton Lee, and Kristina Toutanova. 2019.
\newblock Bert: Pre-training of deep bidirectional transformers for language
  understanding.
\newblock In \emph{NAACL-HLT}.

\bibitem[{Gardner et~al.(2017)Gardner, Grus, Neumann, Tafjord, Dasigi, Liu,
  Peters, Schmitz, and Zettlemoyer}]{Gardner2017AllenNLP}
Matt Gardner, Joel Grus, Mark Neumann, Oyvind Tafjord, Pradeep Dasigi,
  Nelson~F. Liu, Matthew Peters, Michael Schmitz, and Luke~S. Zettlemoyer.
  2017.
\newblock \href {http://arxiv.org/abs/arXiv:1803.07640} {Allennlp: A deep
  semantic natural language processing platform}.
\newblock In \emph{Arxiv}.

\bibitem[{Goyal et~al.(2017)Goyal, Khot, Summers-Stay, Batra, and
  Parikh}]{Goyal2017MakingTV}
Yash Goyal, Tejas Khot, Douglas Summers-Stay, Dhruv Batra, and D.~Parikh. 2017.
\newblock Making the v in vqa matter: Elevating the role of image understanding
  in visual question answering.
\newblock \emph{2017 IEEE Conference on Computer Vision and Pattern Recognition
  (CVPR)}, pages 6325--6334.

\bibitem[{Gupta and Malik(2015)}]{Gupta2015VisualSR}
Saurabh Gupta and Jitendra Malik. 2015.
\newblock Visual semantic role labeling.
\newblock \emph{ArXiv}, abs/1505.04474.

\bibitem[{Heilman and Smith(2009)}]{Heilman2009QuestionGV}
Michael Heilman and Noah~A Smith. 2009.
\newblock Question generation via overgenerating transformations and ranking.
\newblock Technical report, Carnegie-Mellon Univ Pittsburgh pa language
  technologies insT.

\bibitem[{Hudson and Manning(2019)}]{Hudson2019GQAAN}
Drew~A Hudson and Christopher~D Manning. 2019.
\newblock Gqa: A new dataset for real-world visual reasoning and compositional
  question answering.
\newblock \emph{Conference on Computer Vision and Pattern Recognition (CVPR)}.

\bibitem[{Jang et~al.(2017)Jang, Song, Yu, Kim, and Kim}]{Jang2017TGIFQATS}
Y.~Jang, Yale Song, Youngjae Yu, Youngjin Kim, and Gunhee Kim. 2017.
\newblock Tgif-qa: Toward spatio-temporal reasoning in visual question
  answering.
\newblock \emph{2017 IEEE Conference on Computer Vision and Pattern Recognition
  (CVPR)}, pages 1359--1367.

\bibitem[{Jasani et~al.(2019)Jasani, Girdhar, and Ramanan}]{Jasani2019AreWA}
Bhavan Jasani, Rohit Girdhar, and D.~Ramanan. 2019.
\newblock Are we asking the right questions in movieqa?
\newblock \emph{2019 IEEE/CVF International Conference on Computer Vision
  Workshop (ICCVW)}, pages 1879--1882.

\bibitem[{Joshi et~al.(2019)Joshi, Chen, Liu, Weld, Zettlemoyer, and
  Levy}]{Joshi2019SpanBERTIP}
Mandar Joshi, Danqi Chen, Y.~Liu, Daniel~S. Weld, Luke Zettlemoyer, and Omer
  Levy. 2019.
\newblock Spanbert: Improving pre-training by representing and predicting
  spans.
\newblock \emph{Transactions of the Association for Computational Linguistics},
  8:64--77.

\bibitem[{Kingsbury and Palmer(2002)}]{kingsbury2002propbank}
Paul~R Kingsbury and Martha Palmer. 2002.
\newblock From treebank to propbank.
\newblock In \emph{LREC}, pages 1989--1993. Citeseer.

\bibitem[{Krishna et~al.(2017)Krishna, Hata, Ren, Fei-Fei, and
  Niebles}]{Krishna2017DenseCaptioningEI}
R.~Krishna, Kenji Hata, F.~Ren, Li~Fei-Fei, and Juan~Carlos Niebles. 2017.
\newblock Dense-captioning events in videos.
\newblock \emph{2017 IEEE International Conference on Computer Vision (ICCV)},
  pages 706--715.

\bibitem[{Krishna et~al.(2016)Krishna, Zhu, Groth, Johnson, Hata, Kravitz,
  Chen, Kalantidis, Li, Shamma, Bernstein, and Fei-Fei}]{Krishna2016VisualGC}
R.~Krishna, Yuke Zhu, O.~Groth, J.~Johnson, Kenji Hata, J.~Kravitz, Stephanie
  Chen, Yannis Kalantidis, L.~Li, David~A. Shamma, Michael~S. Bernstein, and
  Li~Fei-Fei. 2016.
\newblock Visual genome: Connecting language and vision using crowdsourced
  dense image annotations.
\newblock \emph{International Journal of Computer Vision}, 123:32--73.

\bibitem[{Lee et~al.(2017)Lee, He, Lewis, and Zettlemoyer}]{Lee2017EndtoendNC}
Kenton Lee, Luheng He, M.~Lewis, and Luke Zettlemoyer. 2017.
\newblock End-to-end neural coreference resolution.
\newblock In \emph{EMNLP}.

\bibitem[{Lei et~al.(2018)Lei, Yu, Bansal, and Berg}]{Lei2018TVQALC}
Jie Lei, Licheng Yu, Mohit Bansal, and T.~Berg. 2018.
\newblock Tvqa: Localized, compositional video question answering.
\newblock In \emph{EMNLP}.

\bibitem[{Li et~al.(2020)Li, Zareian, Zeng, Whitehead, Lu, zhong Ji, and
  Chang}]{Li2020CrossmediaSC}
Manling Li, Alireza Zareian, Q.~Zeng, Spencer Whitehead, Di~Lu, Huai zhong Ji,
  and Shih-Fu Chang. 2020.
\newblock Cross-media structured common space for multimedia event extraction.
\newblock In \emph{ACL}.

\bibitem[{Lin(2004)}]{Lin2004ROUGEAP}
Chin-Yew Lin. 2004.
\newblock Rouge: A package for automatic evaluation of summaries.
\newblock In \emph{ACL 2004}.

\bibitem[{Liu et~al.(2019)Liu, Ott, Goyal, Du, Joshi, Chen, Levy, Lewis,
  Zettlemoyer, and Stoyanov}]{Liu2019RoBERTaAR}
Yinhan Liu, Myle Ott, Naman Goyal, Jingfei Du, Mandar Joshi, Danqi Chen, Omer
  Levy, Mike Lewis, Luke Zettlemoyer, and Veselin Stoyanov. 2019.
\newblock Roberta: A robustly optimized bert pretraining approach.
\newblock \emph{arXiv preprint arXiv:1907.11692}.

\bibitem[{Lu et~al.(2019)Lu, Batra, Parikh, and Lee}]{Lu2019ViLBERTPT}
Jiasen Lu, Dhruv Batra, Devi Parikh, and Stefan Lee. 2019.
\newblock \href
  {https://proceedings.neurips.cc/paper/2019/file/c74d97b01eae257e44aa9d5bade97baf-Paper.pdf}
  {Vilbert: Pretraining task-agnostic visiolinguistic representations for
  vision-and-language tasks}.
\newblock In \emph{Advances in Neural Information Processing Systems},
  volume~32. Curran Associates, Inc.

\bibitem[{Maharaj et~al.(2017)Maharaj, Ballas, Rohrbach, Courville, and
  Pal}]{Maharaj2017ADA}
Tegan Maharaj, Nicolas Ballas, Anna Rohrbach, Aaron~C. Courville, and C.~Pal.
  2017.
\newblock A dataset and exploration of models for understanding video data
  through fill-in-the-blank question-answering.
\newblock \emph{2017 IEEE Conference on Computer Vision and Pattern Recognition
  (CVPR)}, pages 7359--7368.

\bibitem[{Miech et~al.(2020)Miech, Alayrac, Smaira, Laptev, Sivic, and
  Zisserman}]{Miech2020EndtoEndLO}
Antoine Miech, Jean-Baptiste Alayrac, Lucas Smaira, I.~Laptev, Josef Sivic, and
  Andrew Zisserman. 2020.
\newblock End-to-end learning of visual representations from uncurated
  instructional videos.
\newblock \emph{2020 IEEE/CVF Conference on Computer Vision and Pattern
  Recognition (CVPR)}, pages 9876--9886.

\bibitem[{Miech et~al.(2019)Miech, Zhukov, Alayrac, Tapaswi, Laptev, and
  Sivic}]{Miech2019HowTo100MLA}
Antoine Miech, D.~Zhukov, Jean-Baptiste Alayrac, Makarand Tapaswi, I.~Laptev,
  and Josef Sivic. 2019.
\newblock Howto100m: Learning a text-video embedding by watching hundred
  million narrated video clips.
\newblock \emph{2019 IEEE/CVF International Conference on Computer Vision
  (ICCV)}, pages 2630--2640.

\bibitem[{Ott et~al.(2019)Ott, Edunov, Baevski, Fan, Gross, Ng, Grangier, and
  Auli}]{ott2019fairseq}
Myle Ott, Sergey Edunov, Alexei Baevski, Angela Fan, Sam Gross, Nathan Ng,
  David Grangier, and Michael Auli. 2019.
\newblock fairseq: A fast, extensible toolkit for sequence modeling.
\newblock In \emph{Proceedings of NAACL-HLT 2019: Demonstrations}.

\bibitem[{Papineni et~al.(2002)Papineni, Roukos, Ward, and
  Zhu}]{Papineni2002BleuAM}
Kishore Papineni, S.~Roukos, T.~Ward, and Wei-Jing Zhu. 2002.
\newblock Bleu: a method for automatic evaluation of machine translation.
\newblock In \emph{ACL}.

\bibitem[{Paszke et~al.(2019)Paszke, Gross, Massa, Lerer, Bradbury, Chanan,
  Killeen, Lin, Gimelshein, Antiga, Desmaison, Kopf, Yang, DeVito, Raison,
  Tejani, Chilamkurthy, Steiner, Fang, Bai, and Chintala}]{NEURIPS2019_9015}
Adam Paszke, Sam Gross, Francisco Massa, Adam Lerer, James Bradbury, Gregory
  Chanan, Trevor Killeen, Zeming Lin, Natalia Gimelshein, Luca Antiga, Alban
  Desmaison, Andreas Kopf, Edward Yang, Zachary DeVito, Martin Raison, Alykhan
  Tejani, Sasank Chilamkurthy, Benoit Steiner, Lu~Fang, Junjie Bai, and Soumith
  Chintala. 2019.
\newblock \href
  {http://papers.neurips.cc/paper/9015-pytorch-an-imperative-style-high-performance-deep-learning-library.pdf}
  {Pytorch: An imperative style, high-performance deep learning library}.
\newblock In H.~Wallach, H.~Larochelle, A.~Beygelzimer, F.~d~Alch\'{e}-Buc,
  E.~Fox, and R.~Garnett, editors, \emph{Advances in Neural Information
  Processing Systems 32}, pages 8024--8035. Curran Associates, Inc.

\bibitem[{Pennington et~al.(2014)Pennington, Socher, and
  Manning}]{Pennington2014GloveGV}
Jeffrey Pennington, R.~Socher, and Christopher~D. Manning. 2014.
\newblock Glove: Global vectors for word representation.
\newblock In \emph{EMNLP}.

\bibitem[{Pradhan et~al.(2013)Pradhan, Moschitti, Xue, Ng, Bj{\"o}rkelund,
  Uryupina, Zhang, and Zhong}]{pradhan2013towards}
Sameer Pradhan, Alessandro Moschitti, Nianwen Xue, Hwee~Tou Ng, Anders
  Bj{\"o}rkelund, Olga Uryupina, Yuchen Zhang, and Zhi Zhong. 2013.
\newblock Towards robust linguistic analysis using ontonotes.
\newblock In \emph{Proceedings of the Seventeenth Conference on Computational
  Natural Language Learning}, pages 143--152.

\bibitem[{Ren et~al.(2015)Ren, He, Girshick, and Sun}]{Ren2015FasterRT}
Shaoqing Ren, Kaiming He, Ross~B. Girshick, and J.~Sun. 2015.
\newblock Faster r-cnn: Towards real-time object detection with region proposal
  networks.
\newblock \emph{IEEE Transactions on Pattern Analysis and Machine
  Intelligence}, 39:1137--1149.

\bibitem[{Sadhu et~al.(2020)Sadhu, Chen, and Nevatia}]{Sadhu2020VideoOG}
Arka Sadhu, K.~Chen, and R.~Nevatia. 2020.
\newblock Video object grounding using semantic roles in language description.
\newblock \emph{2020 IEEE/CVF Conference on Computer Vision and Pattern
  Recognition (CVPR)}, pages 10414--10424.

\bibitem[{Shi and Lin(2019)}]{shi2019simple}
Peng Shi and Jimmy Lin. 2019.
\newblock Simple bert models for relation extraction and semantic role
  labeling.
\newblock \emph{arXiv preprint arXiv:1904.05255}.

\bibitem[{Sigurdsson et~al.(2016)Sigurdsson, Varol, Wang, Farhadi, Laptev, and
  Gupta}]{Sigurdsson2016HollywoodIH}
Gunnar~A. Sigurdsson, G{\"u}l Varol, X.~Wang, Ali Farhadi, I.~Laptev, and
  A.~Gupta. 2016.
\newblock Hollywood in homes: Crowdsourcing data collection for activity
  understanding.
\newblock In \emph{ECCV}.

\bibitem[{Silberer and Pinkal(2018)}]{Silberer2018GroundingSR}
Carina Silberer and Manfred Pinkal. 2018.
\newblock Grounding semantic roles in images.
\newblock In \emph{EMNLP}.

\bibitem[{Strubell et~al.(2018)Strubell, Verga, Andor, Weiss, and
  McCallum}]{Strubell2018LinguisticallyInformedSF}
Emma Strubell, Patrick Verga, Daniel Andor, David Weiss, and Andrew McCallum.
  2018.
\newblock \href {https://doi.org/10.18653/v1/D18-1548} {Linguistically-informed
  self-attention for semantic role labeling}.
\newblock In \emph{Proceedings of the 2018 Conference on Empirical Methods in
  Natural Language Processing}, pages 5027--5038, Brussels, Belgium.
  Association for Computational Linguistics.

\bibitem[{Tapaswi et~al.(2016)Tapaswi, Zhu, Stiefelhagen, Torralba, Urtasun,
  and Fidler}]{Tapaswi2016MovieQAUS}
Makarand Tapaswi, Y.~Zhu, R.~Stiefelhagen, A.~Torralba, R.~Urtasun, and
  S.~Fidler. 2016.
\newblock Movieqa: Understanding stories in movies through question-answering.
\newblock \emph{2016 IEEE Conference on Computer Vision and Pattern Recognition
  (CVPR)}, pages 4631--4640.

\bibitem[{Vaswani et~al.(2017)Vaswani, Shazeer, Parmar, Uszkoreit, Jones,
  Gomez, Kaiser, and Polosukhin}]{Vaswani2017AttentionIA}
Ashish Vaswani, Noam Shazeer, Niki Parmar, Jakob Uszkoreit, Llion Jones,
  Aidan~N Gomez, \L~ukasz Kaiser, and Illia Polosukhin. 2017.
\newblock \href
  {https://proceedings.neurips.cc/paper/2017/file/3f5ee243547dee91fbd053c1c4a845aa-Paper.pdf}
  {Attention is all you need}.
\newblock In \emph{Advances in Neural Information Processing Systems},
  volume~30. Curran Associates, Inc.

\bibitem[{Vedantam et~al.(2015)Vedantam, Zitnick, and
  Parikh}]{Vedantam2015CIDErCI}
Ramakrishna Vedantam, C.~L. Zitnick, and D.~Parikh. 2015.
\newblock Cider: Consensus-based image description evaluation.
\newblock \emph{2015 IEEE Conference on Computer Vision and Pattern Recognition
  (CVPR)}, pages 4566--4575.

\bibitem[{Wang et~al.(2016)Wang, Xiong, Wang, Qiao, Lin, Tang, and
  Gool}]{Wang2016TemporalSN}
L.~Wang, Yuanjun Xiong, Zhe Wang, Yu~Qiao, D.~Lin, X.~Tang, and L.~Gool. 2016.
\newblock Temporal segment networks: Towards good practices for deep action
  recognition.
\newblock In \emph{ECCV}.

\bibitem[{Xie et~al.(2018)Xie, Sun, Huang, Tu, and
  Murphy}]{Xie2018RethinkingSF}
Saining Xie, C.~Sun, J.~Huang, Zhuowen Tu, and Kevin Murphy. 2018.
\newblock Rethinking spatiotemporal feature learning: Speed-accuracy trade-offs
  in video classification.
\newblock In \emph{ECCV}.

\bibitem[{Xu et~al.(2017)Xu, Zhao, Xiao, Wu, Zhang, He, and
  Zhuang}]{xu2017video}
Dejing Xu, Zhou Zhao, Jun Xiao, Fei Wu, Hanwang Zhang, Xiangnan He, and Yueting
  Zhuang. 2017.
\newblock Video question answering via gradually refined attention over
  appearance and motion.
\newblock In \emph{ACM Multimedia}.

\bibitem[{Yang et~al.(2020)Yang, Zhu, Wang, Yi, Zadeh, and
  Morency}]{Yang2020WhatGT}
Jianing Yang, Yuying Zhu, Yongxin Wang, Ruitao Yi, Amir Zadeh, and
  Louis-Philippe Morency. 2020.
\newblock \href {http://arxiv.org/abs/2007.03626} {What gives the answer away?
  question answering bias analysis on video qa datasets}.
\newblock \emph{arXiv:2007.03626 [cs, stat]}.

\bibitem[{Yatskar et~al.(2016)Yatskar, Zettlemoyer, and Farhadi}]{yatskar2016}
Mark Yatskar, Luke Zettlemoyer, and Ali Farhadi. 2016.
\newblock Situation recognition: Visual semantic role labeling for image
  understanding.
\newblock In \emph{Conference on Computer Vision and Pattern Recognition}.

\bibitem[{Yu et~al.(2019)Yu, Xu, Yu, Yu, Zhao, Zhuang, and
  Tao}]{yu2019activityqa}
Zhou Yu, Dejing Xu, Jun Yu, Ting Yu, Zhou Zhao, Yueting Zhuang, and Dacheng
  Tao. 2019.
\newblock Activitynet-qa: A dataset for understanding complex web videos via
  question answering.
\newblock In \emph{AAAI}, pages 9127--9134.

\bibitem[{Zellers et~al.(2019)Zellers, Bisk, Farhadi, and
  Choi}]{Zellers2019FromRT}
Rowan Zellers, Yonatan Bisk, Ali Farhadi, and Yejin Choi. 2019.
\newblock From recognition to cognition: Visual commonsense reasoning.
\newblock \emph{2019 IEEE/CVF Conference on Computer Vision and Pattern
  Recognition (CVPR)}, pages 6713--6724.

\bibitem[{Zeng et~al.(2017)Zeng, Chen, Chuang, Liao, Niebles, and
  Sun}]{Zeng2017LeveragingVD}
Kuo-Hao Zeng, Tseng-Hung Chen, Ching-Yao Chuang, Yuan-Hong Liao, Juan~Carlos
  Niebles, and Min Sun. 2017.
\newblock Leveraging video descriptions to learn video question answering.
\newblock In \emph{AAAI}.

\bibitem[{Zhang* et~al.(2020)Zhang*, Kishore*, Wu*, Weinberger, and
  Artzi}]{Zhang2020BERTScoreET}
Tianyi Zhang*, Varsha Kishore*, Felix Wu*, Kilian~Q. Weinberger, and Yoav
  Artzi. 2020.
\newblock \href {https://openreview.net/forum?id=SkeHuCVFDr} {Bertscore:
  Evaluating text generation with bert}.
\newblock In \emph{International Conference on Learning Representations}.

\bibitem[{Zhu and Yang(2020)}]{Zhu2020ActBERTLG}
Linchao Zhu and Yi~Yang. 2020.
\newblock Actbert: Learning global-local video-text representations.
\newblock In \emph{Proceedings of the IEEE/CVF Conference on Computer Vision
  and Pattern Recognition (CVPR)}.

\end{thebibliography}
\bibliographystyle{acl_natbib}
\clearpage
\newpage

\appendix
\begin{center}
  {\large \bf Appendix \par}
\end{center}

\setcounter{table}{0}
\setcounter{figure}{0}
\setcounter{equation}{0}

This is the appendix for the paper ``Video Question Answering with Phrases via Semantic Roles''. 
The appendix provides details on
\begin{enumerate}
    \itemsep0em
    \item Dataset construction and Dataset statistics (Section \ref{sec:suppl_dscons})
    \item Implementation Details for both the Metrics as well as the Models (Section \ref{sec:suppl_impl_details}).
    \item Visualization of Model Outputs (Section \ref{sec:suppl_mdl_out_vis})
    \item Code and Data are publicly available \footnote{\gh{}}.
\end{enumerate}

\section{Dataset Construction}
\label{sec:suppl_dscons}
We first discuss semantic-role labeling used in natural language processing. 
Then, we detail the dataset construction process used for \anetdsn{} and \chdsn{} (Section \ref{ss:suppl_ds_cons_proc}) and then provide the dataset statistics (Section \ref{ss:suppl_ds_stats}).

\subsection{Semantic Role Labeling}
\label{ss:suppl_srl_disc}
Semantic-Role Labels extract out high-level meanings from a natural language description.
Two widely used SRL annotations are PropBank \cite{kingsbury2002propbank} and FrameNet \cite{baker1998framenet}.
Here we use SRLs which follow PropBank annotation guidelines (see \cite{bonial2012propbank} for complete guideline).

Most commonly used argument roles are 
\begin{itemize}
    \itemsep0em
    \item \txi{V}: the verb. All remaining roles are dependent on this verb. While the numbered arguments differ slightly based on the verb used, they share common themes across verbs as listed below (see \cite{bonial2012propbank} for full details). 
    For instance, ``cut'' is a Verb.
    \item \txi{ARG0}: the agent, or the one causing the verb. For most action verbs, this is usually a human or an animal.
    For instance, ``A person cuts a vegetable'', ``A person'' is \txi{ARG0}.
    \item \txi{ARG1}: the object, on which the action is being performed.
    In ``A person cuts a vegetable'', ``a vegetable'' is \txi{ARG1}.
    \item \txi{ARG2}: the tool being used for the verb, or someone who benefits from the verb. For instance, in ``A person is cutting a vegetable with a knife'', ``with a knife'' denotes the tool and is \txi{ARG2}. 
    In ``A person throws a basketball to the basket'', ``to the basket'' denotes the benefactor and is \txi{ARG2}.
    \item \txi{ARGM-LOC} or simply \txi{LOC} denotes the place or location where the verb takes place. For instance, in ``A person is cutting a vegetable on a plate'', ``on a plate'' is the \txi{LOC}.
\end{itemize}

To assign SRLs to language descriptions we use allennlp library \cite{Gardner2017AllenNLP} which provides an implementation of a BERT \cite{Devlin2019BERTPO} based semantic-role labeler \cite{shi2019simple}.
The system achieves $86.49$ F1 score on OntoNotes \cite{pradhan2013towards} 5.0 dataset.

\subsection{Construction Process}
\label{ss:suppl_ds_cons_proc}
Both \anetdsn{} and \chdsn{} follow the same process with few subtle differences:
\begin{enumerate}
    \itemsep0em
    \item Pre-Process Data:
    \begin{itemize}
        \item Assign semantic role labels (SRLs) to video descriptions using SRL labeller \cite{shi2019simple}.
        \item Remove stopword verbs with lemmas: ``be'', ``start'', ``end'', ``begin'', ``stop'', ``lead'', ``demonstrate'', ``do''.
        \item For the original descriptions spread across multiple video segments, combine the sentences into a document. Use a co-reference resolution model on this model (we use \cite{Lee2017EndtoendNC} with SpanBERT embeddings \cite{Joshi2019SpanBERTIP} provided in allennlp library \cite{Gardner2017AllenNLP}).
        \item Replace the following pronouns: ``they'', ``he'', ``she'', ``his'', ``her'', ``it'' with the relevant noun-phrase obtained from the co-reference resolution output.
    \end{itemize}
    \item Query-Generation:
    \begin{itemize}
        \item For each verb-role set within a description (each description can have multiple verbs), consider the role set \txi{ARG0, ARG1, V, ARG2, LOC} for \anetdsn{} and \txi{ARG1, V, ARG2, LOC} for \chdsn{}. 
        \item If there are at least $3$ verb-roles for the given verb, for each SRL replace it with a query token (with ${<}\texttt{Q}{-}\{\texttt{R}\}{>}$ where $R$ is the role). This forms one query. Repeat for all SRLs in the considered set. 
        \item The minimum of $3$ verb-roles is present to avoid ambiguity in the query. Limiting the argument role-set helps in generating queries less likely to have strong language-priors (though as seen in qualitative examples, some priors are still present).
        \item After the queries are generated, create lemmatized verbs, and nouns set for each query, and store the video segment ids in a dictionary. This is similar to the process used in \cite{Sadhu2020VideoOG}, with the difference that we additionally have query-tokens.
        \item For each query, use the dictionary to sample set of video segment ids which share the same semantic role structure, but for the query-token have a different answer. These are used for matching when computing the scores for the validation and testing set using the contrastive score.
    \end{itemize}
    \item Creating Train/Test Splits:
    \begin{itemize}
        \item Keep the training set for each dataset the same.
        \item For validation and testing, we split the dataset based on the video ids (half video ids are set as validation, and half as testing). The queries are then split based on the video ids.
        \item Note that while contrastive sampling is done before validation test split. So validation and test ids are used for computing the other's score for contrastive sampling. This is similar to the setting used in \cite{Sadhu2020VideoOG} as the total number of videos available for validation, and testing are insufficient for contrastive sampling.
    \end{itemize}
\end{enumerate}

\subsection{Dataset Statistics}
\label{ss:suppl_ds_stats}
\begin{table*}[!ht]
\centering
\begin{tabular}{@{}c|c|ccc|ccc@{}}
\toprule
           &               & \multicolumn{3}{c|}{\anetdsn{}} & \multicolumn{3}{c}{\chdsn{}} \\ \midrule
           &               & Train      & Val  & Test & Train    & Val  & Test \\
Overall    & Videos        & 30337  & 2729 & 2739 & 7733  & 860  & 876  \\
           & Queries       & 147439 & 7414 & 7238 & 59329 & 4431 & 4520 \\
           & Query Length  & 8.03   & 6.03 & 6    & 7.11  & 5.6  & 5.62 \\
           & Answer Length & 2.2    & 2.33 & 2.33 & 1.83  & 1.96 & 1.94 \\
           & Vocabulary    & 4597   & 3261 & 3261 & 1479  & 884  & 884  \\
           &               &        &      &      &       &      &      \\
\txi{ARG0} & Videos        & 24483  & 1372 & 1419 &       &      &      \\
           & Queries       & 37218  & 1603 & 1643 &       &      &      \\
           & Query Length  & 7.31   & 5.73 & 5.65 &       &      &      \\
           & Answer Length & 2.51   & 2.37 & 2.48 &       &      &      \\
           & Vocabulary    & 1763   & 840  & 840  &       &      &      \\
           &               &        &      &      &       &      &      \\
\txi{V}    & Videos        & 29922  & 1737 & 1733 & 7733  & 802  & 811  \\
           & Queries       & 52447  & 2247 & 2187 & 27745 & 1824 & 1829 \\
           & Query Length  & 9.2    & 7.26 & 7.18 & 7.7   & 6.37 & 6.44 \\
           & Answer Length & 1      & 1    & 1    & 1     & 1    & 1    \\
           & Vocabulary    & 1860   & 1167 & 1167 & 678   & 377  & 377  \\
           &               &        &      &      &       &      &      \\
\txi{ARG1} & Videos        & 24863  & 1810 & 1793 & 7600  & 808  & 828  \\
           & Queries       & 36787  & 2250 & 2179 & 21557 & 1857 & 1874 \\
           & Query Length  & 7.4    & 5.4  & 5.43 & 6.43  & 5.07 & 5.04 \\
           & Answer Length & 2.8    & 2.82 & 2.83 & 2.31  & 2.39 & 2.39 \\
           & Vocabulary    & 3560   & 2124 & 2124 & 935   & 527  & 527  \\
           &               &        &      &      &       &      &      \\
\txi{ARG2} & Videos        & 12048  & 850  & 805  & 5433  & 490  & 522  \\
           & Queries       & 14321  & 941  & 886  & 8279  & 651  & 699  \\
           & Query Length  & 7.49   & 5.45 & 5.36 & 6.94  & 5.13 & 5.13 \\
           & Answer Length & 3.55   & 3.69 & 3.62 & 3.11  & 3.22 & 3.04 \\
           & Vocabulary    & 2607   & 1326 & 1326 & 556   & 365  & 365  \\
           &               &        &      &      &       &      &      \\
\txi{LOC}  & Videos        & 6025   & 340  & 319  & 1578  & 87   & 112  \\
           & Queries       & 6666   & 373  & 343  & 1748  & 99   & 118  \\
           & Query Length  & 7.57   & 5.17 & 5.35 & 6.93  & 4.75 & 5.06 \\
           & Answer Length & 3.61   & 3.87 & 3.63 & 3.22  & 3.19 & 3.08 \\
           & Vocabulary    & 1390   & 669  & 669  & 265   & 138  & 138  \\ \bottomrule
\end{tabular}
\caption{Detailed dataset statistics for both \anetdsn{} and \chdsn{} with respect to different argument roles. Recall that \txi{ARG0} is not present in \chdsn{}, and hence the corresponding rows are kept blank.}
\label{tab:suppl_ds_stats}
\end{table*}

\begin{figure}[!ht]
    \centering
    \includegraphics[width=\linewidth]{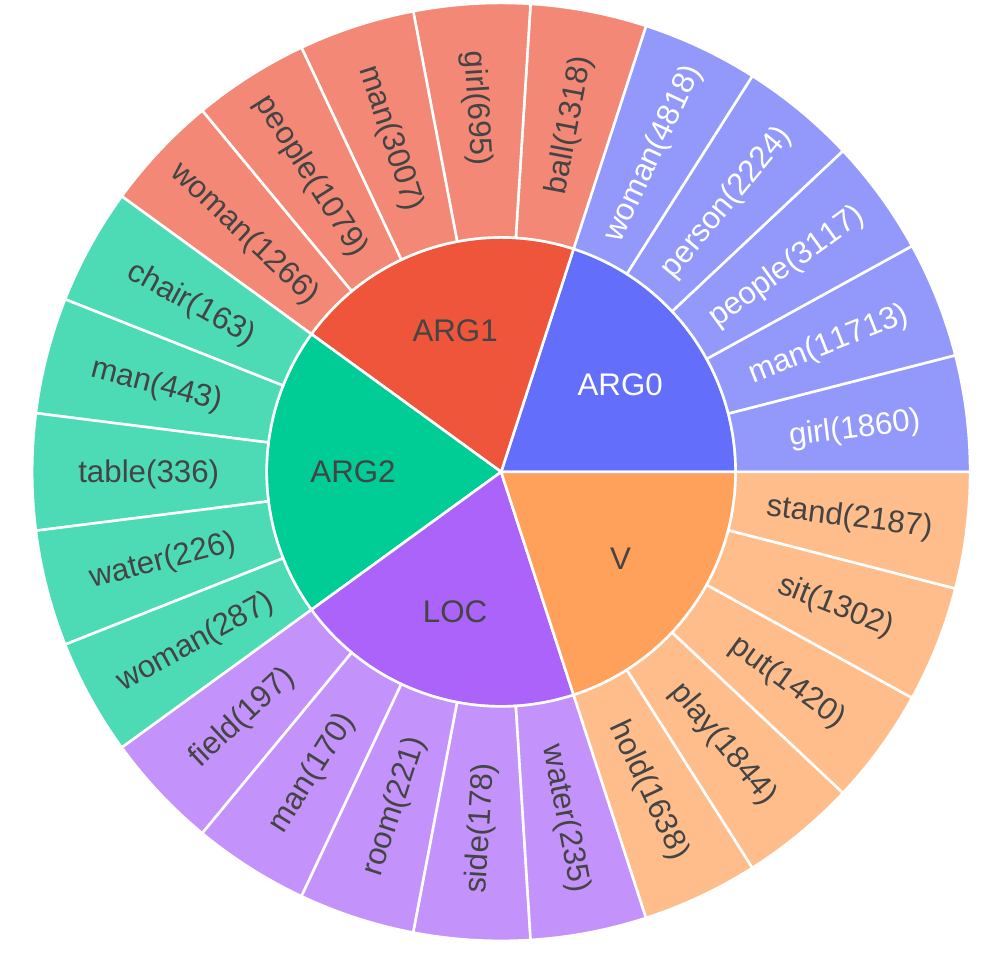}
    (a) Top-5 lemmatized nouns or verbs for the considered semantic roles in \anetdsn{}
    \includegraphics[width=\linewidth]{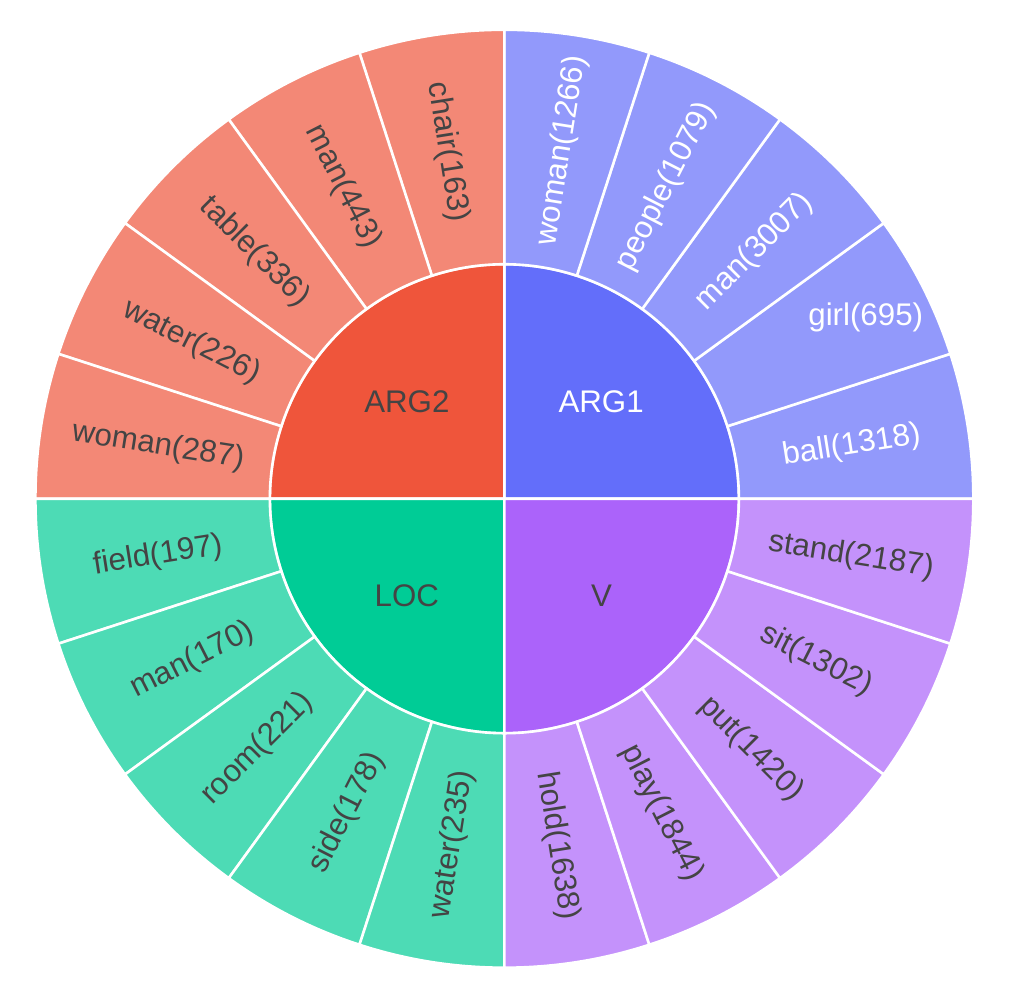}
    (b) Top-5 lemmatized nouns or verbs for the considered semantic roles in \chdsn{}
    \caption{Lemma Distribution for both \anetdsn{} and \chdsn{}. The number of instances across the whole dataset are given in the parenthesis of each lemmatized noun or verb.}
    \label{fig:suppl_ds_stats}
\end{figure}
Dataset statistics can be found in Table \ref{tab:suppl_ds_stats}.
Lemma distributions are visualized in Figure \ref{fig:suppl_ds_stats}
Overall, we find slightly skewed distribution of Argument roles across the datasets. For instance, \txi{ARG0, ARG1} are much more frequent than \txi{ARG2} and \txi{LOC}.
Also, since every SRL needs to have a verb (\txi{V}), the distribution of the videos is the same as the overall.

As shown in Table \ref{tab:suppl_ds_stats}, vocabularies in both the train and validation/test sets for each argument role (slot) are reasonably large compared (eg. 60\% for ARG1) to the total vocabulary and not too limited. This results is further consistent across both datasets.
\section{Implementation Details}
\label{sec:suppl_impl_details}

We first report the implementation details for the metrics (Section \ref{ss:suppl_metr_impl}).
Then, we detail the model implementation details (Section \ref{ss:suppl_model_impl}).

\subsection{Metric Implementation}
\label{ss:suppl_metr_impl}
For Bleu \cite{Papineni2002BleuAM}, Rouge \cite{Lin2004ROUGEAP}, Meteor \cite{Banerjee2005METEORAA}, and CIDEr \cite{Vedantam2015CIDErCI} we use the implementations provided in coco-captions repository\footnote{github url: https://github.com/tylin/coco-caption} \cite{Chen2015MicrosoftCC}.

For BERTScore we use the official implementation \footnote{github url: https://github.com/Tiiiger/bert\_score}

\txbu{BLEU-2:} computes Bleu with n-gram with $n{=}2$. We use sentence-bleu score instead of the more commonly used corpus bleu score. This is further used for contrastive sampling.

\txbu{ROUGE:} we use ROUGE-L which computes the longest common sub-sequence. 

\txbu{METEOR:} we use Meteor 1.5 version \cite{denkowski:lavie:meteor-wmt:2014}.

\txbu{CIDEr:} we use CIDEr-D implementation which includes idf-weighting.

\txbu{BertScore:} we use BertScore with hash ``roberta-large\_L17\_idf\_version=0.3.5(hug\_trans=3.0.2)-rescaled''

We show examples of computing the metrics.

\subsection{Model Implementation}
\label{ss:suppl_model_impl}

We report all model implementation details. 

\textbf{General Settings:} Our code is implemented using Pytorch \cite{NEURIPS2019_9015}. For Transformer, we use the implementation provided in FairSeq \cite{ott2019fairseq}.
The vocabulary consists of $5$k words for \anetdsn{} and $3$k words for \chdsn{}.
The segment features are of dimension $3072$ and $512$ for \anetdsn{} and \chdsn{} respectively obtained from TSN \cite{Wang2016TemporalSN} and S3D \cite{Xie2018RethinkingSF} trained on HowTo100M \cite{Miech2019HowTo100MLA} using the loss function presented in \cite{Miech2020EndtoEndLO}
\footnote{https://github.com/antoine77340/S3D\_HowTo100M}.
The proposal features are of dimension $1024$ and only used for \anetdsn{} extracted using FasterRCNN \cite{Ren2015FasterRT} trained on Visual Genome \cite{Krishna2016VisualGC}.

For all cases, we report the output dimension of MLP. 
Unless otherwise stated, MLP is followed by ReLU activation.

\textbf{Decoder:} 
The decoder uses an input of $T \times 512$ (where $T$ refers to the length of the input embedding). 
Note that for \tnamem{Lang}, $T$ is same as sequence length of the query, for \tnamem{BUTD} $T{=}1$, for \tnamem{VOG}, $T$ is number of SRLs $*$  number of segment features. 
For \tnamem{MTX}, $T$ is sequence length of query $+$ number of segment features.
To generate output sequences, we use the usual beam-search with a beam-size of $2$, with a temperature of $1.0$.

\textbf{Encoder:}
Encoder differs based on the specific model. 
All encoders are transformer based using $8$ attention heads and $3$ layers unless otherwise mentioned.
 
\textbf{\tnamem{Lang}}: 
The language encoder uses $3$ encoding layers, with $8$ attention heads each.
The embedding layer uses a dimension of $512$.

\textbf{\tnamem{BUTD}}: We use the same language query, with and pre-pend a $[CLS]$ token. The embedding of the $[CLS]$ token serves as the language embedding, and is passed through a MLP of dimension $512$.
The language encoder is the same as \tnamem{Lang}.
The segment features are passed through MLP of dimension $512$.
If proposal features are used, they are passed through a separate MLP of dimension $512$.
The language embedding (also of dimension $512$) is used to compute attention score with the visual features, and finally obtain an attended visual feature.
These attended visual features are concatenated with the language embedding along the last axis, and then passed to the decoder. 

\textbf{\tnamem{VOG}}: We use the same language encoder, but further use the SRL phrase start and end-points for the phrase encoder.
The phrase encoder uses these start and end points to gather the language embeddings corresponding to these start and end points, concatenate them (dimension $512{+}512{=}1024$) and use MLP with dimension $512$. 
This gives an output of the phrase encoder of size number of SRLs $*s 512$.
The phrase encoded query is then concatenated with all the segment features and passed through a MLP. 
Finally a multi-modal transformer encoder is applied over the phrase encoded input, and is passed to the language decoder.

\textbf{\tnamem{MTX}}:
We collate all the language tokens (passed through embedding layer) as well as segment features passed through MLP, to get all features of dimension $512$.
A transformer based encoder is applied on these features, and the output is passed to the decoder.

\textbf{Training}:
We train using standard cross-entropy loss.
The decoder is trained using teacher forcing.
All models are trained for $10$ epochs with batch size of $32$. 
On a TitanX, for \anetdsn{} each epoch takes around $30-40$ mins.
Our training infrastructure included a 8 GPU Titan X machine

\section{Visualization}
\label{sec:suppl_mdl_out_vis}

\begin{figure*}
    \centering
    \includegraphics[width=\linewidth]{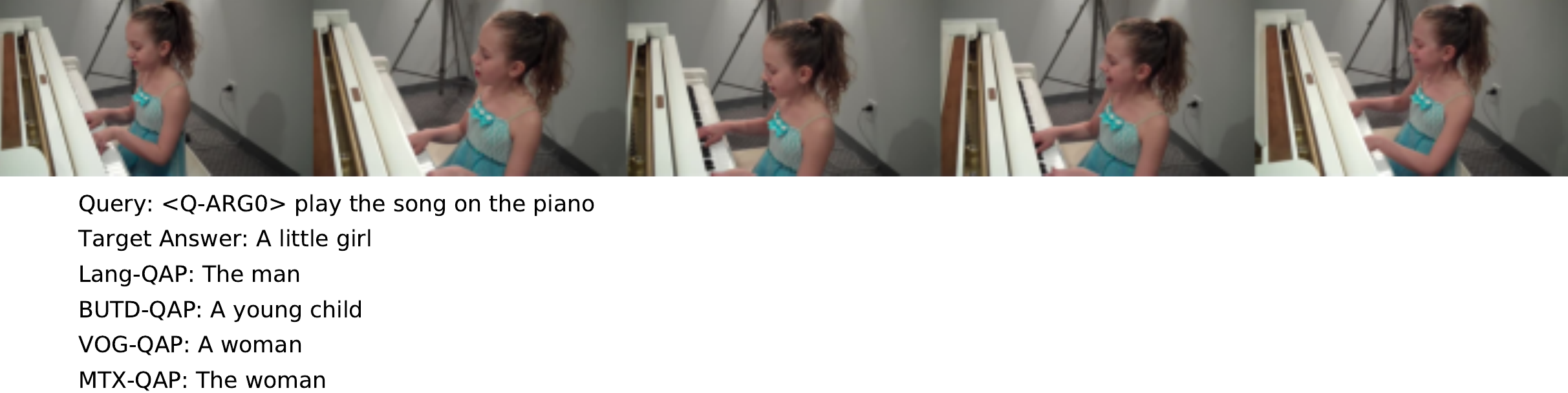}
    \includegraphics[width=\linewidth]{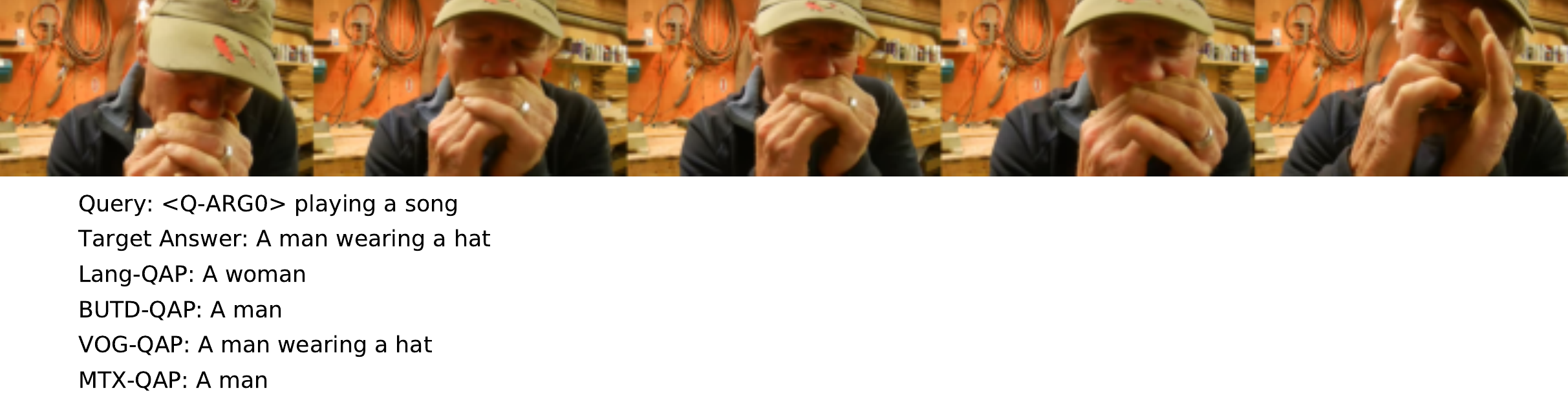}
    (a) Query of type \txi{ARG0}
    \bigskip
    \bigskip
    
    \includegraphics[width=\linewidth]{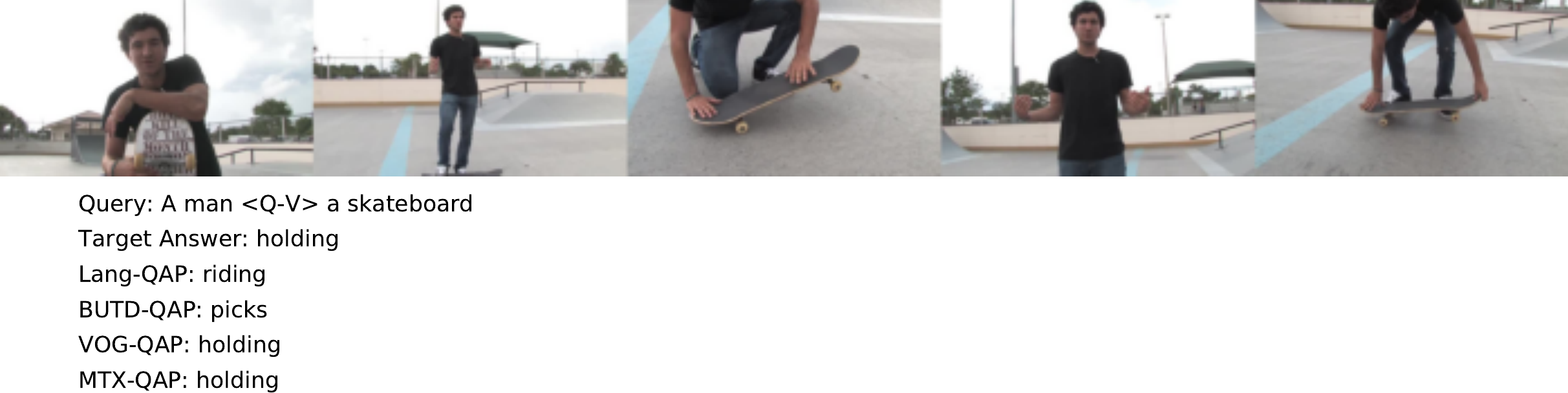}
    \includegraphics[width=\linewidth]{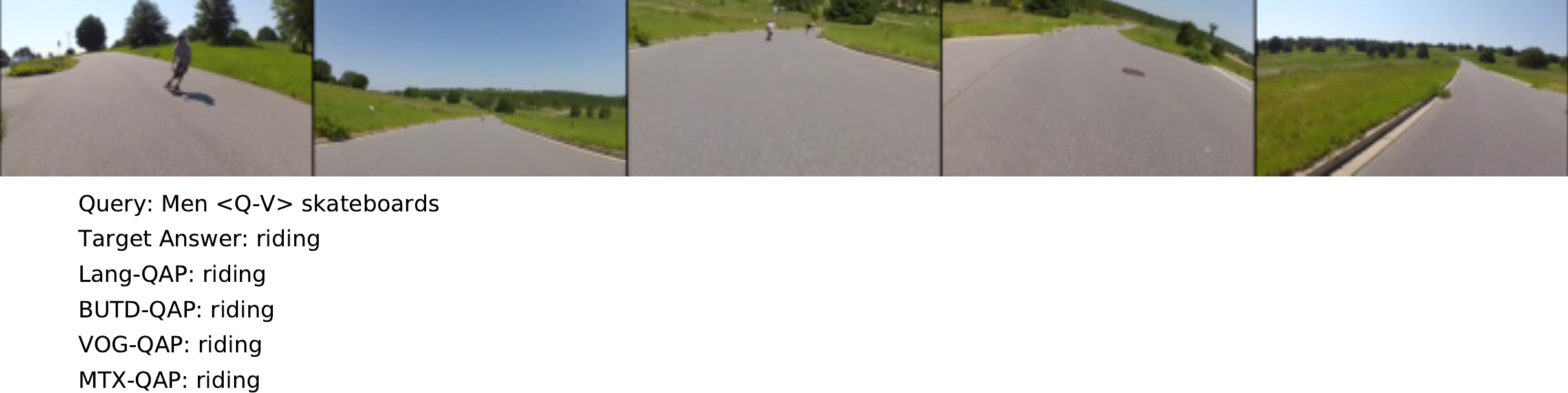}
    (b) Query of type \txi{V}
    
    \caption{Queries of Type \txi{ARG0} and \txi{V} on \anetdsn{}}
    \label{fig:suppl_a0_v_anet_vis}

\end{figure*}

\begin{figure*}
    \centering
    \includegraphics[width=0.9\linewidth]{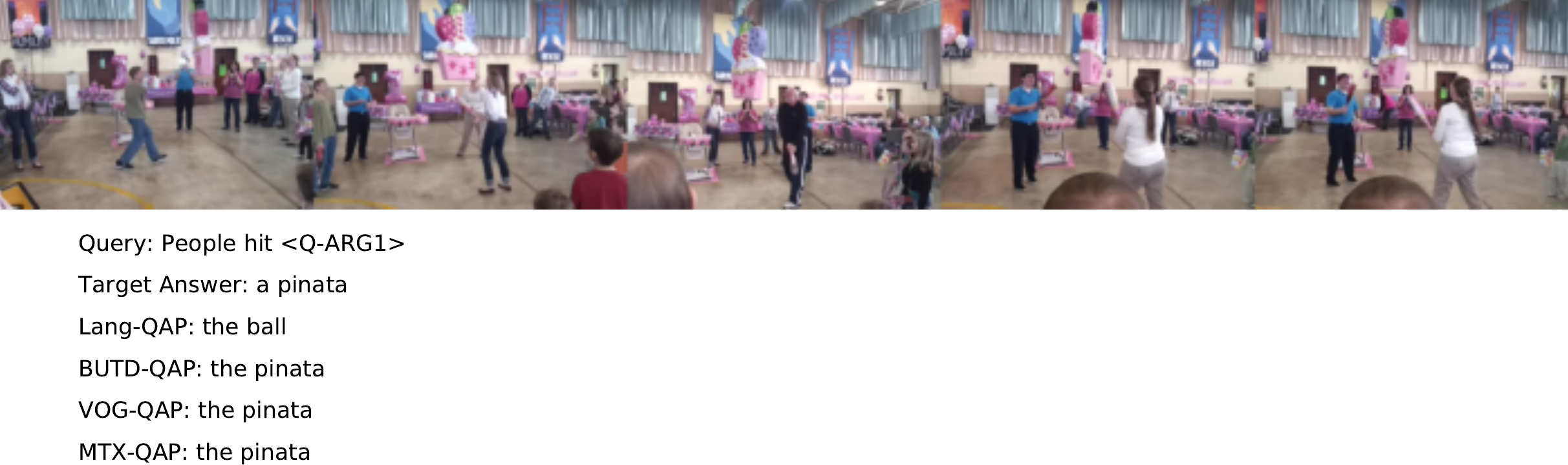}
    \includegraphics[width=0.9\linewidth]{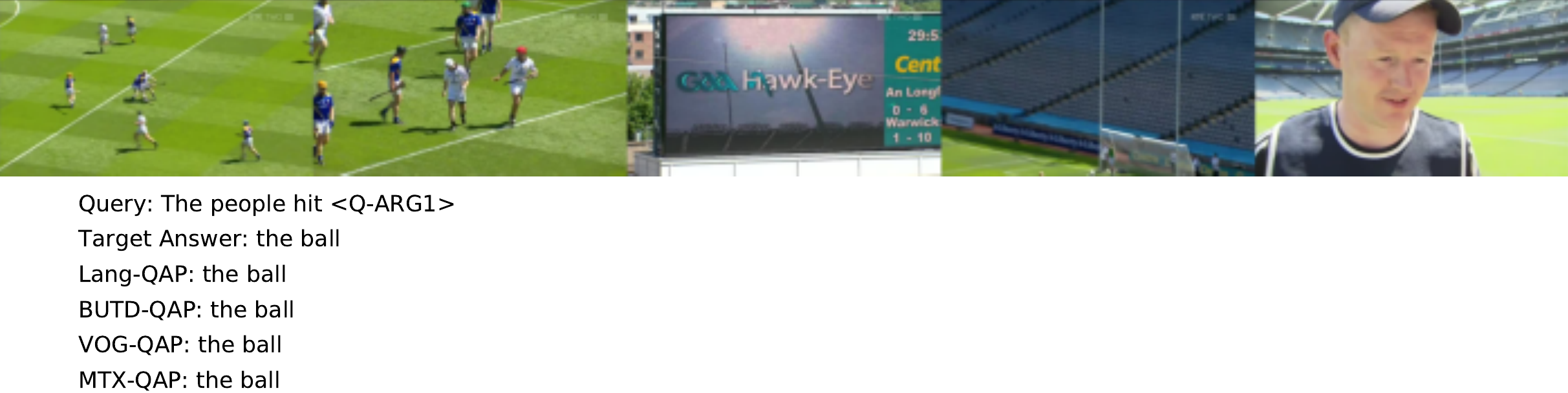}\\
    (a) Query of type \txi{ARG1}
    \bigskip
    \bigskip
    
    \includegraphics[width=0.9\linewidth]{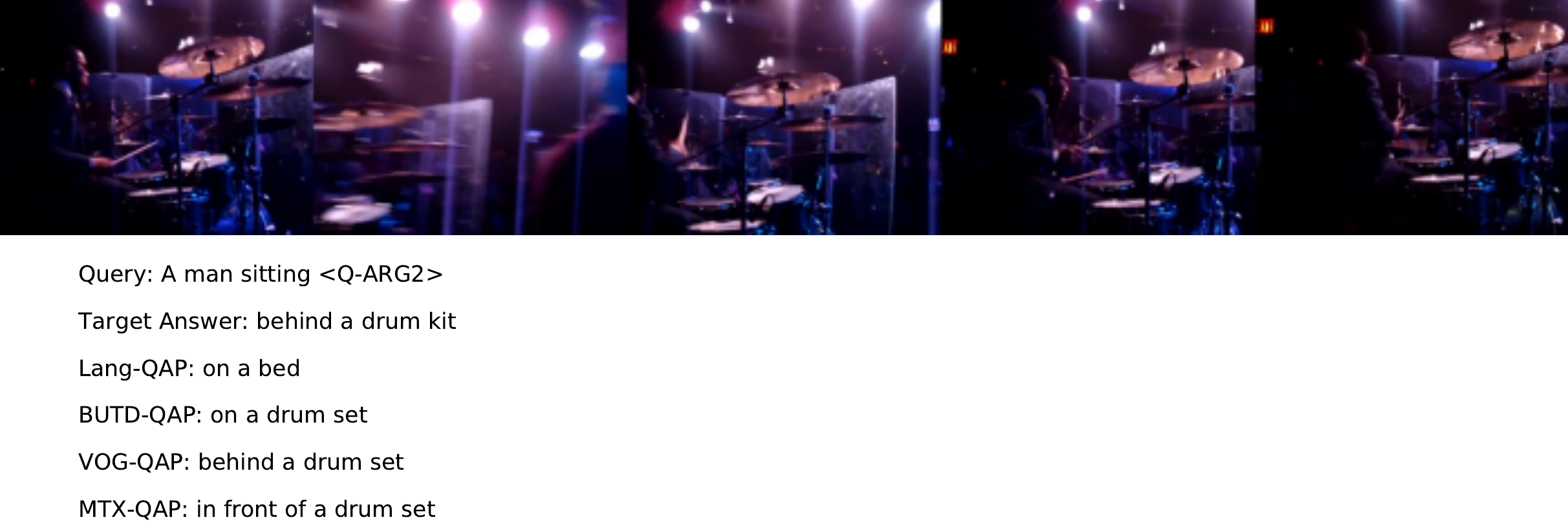}
    \includegraphics[width=0.9\linewidth]{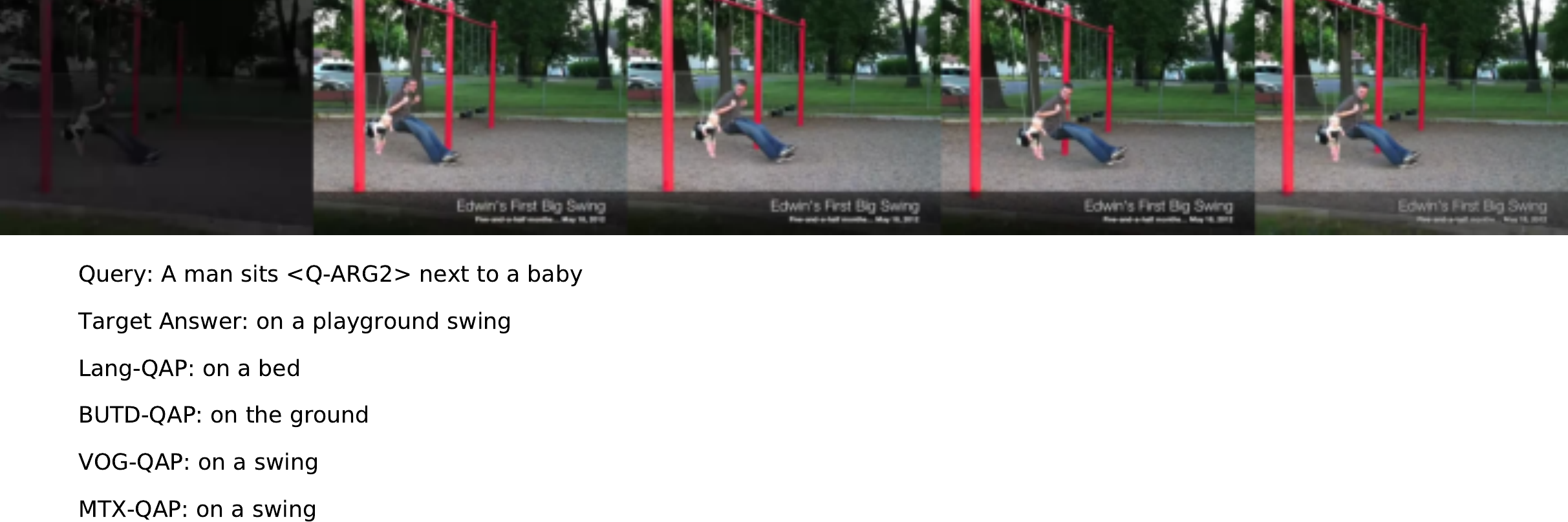}\\
    (b) Query of type \txi{ARG2}
    
    \caption{Queries of Type \txi{ARG1} and \txi{ARG2} on \anetdsn{}}
    \label{fig:suppl_a1_a2_anet_vis}

\end{figure*}

\begin{figure*}[!ht]
    \centering
    \includegraphics[width=\linewidth]{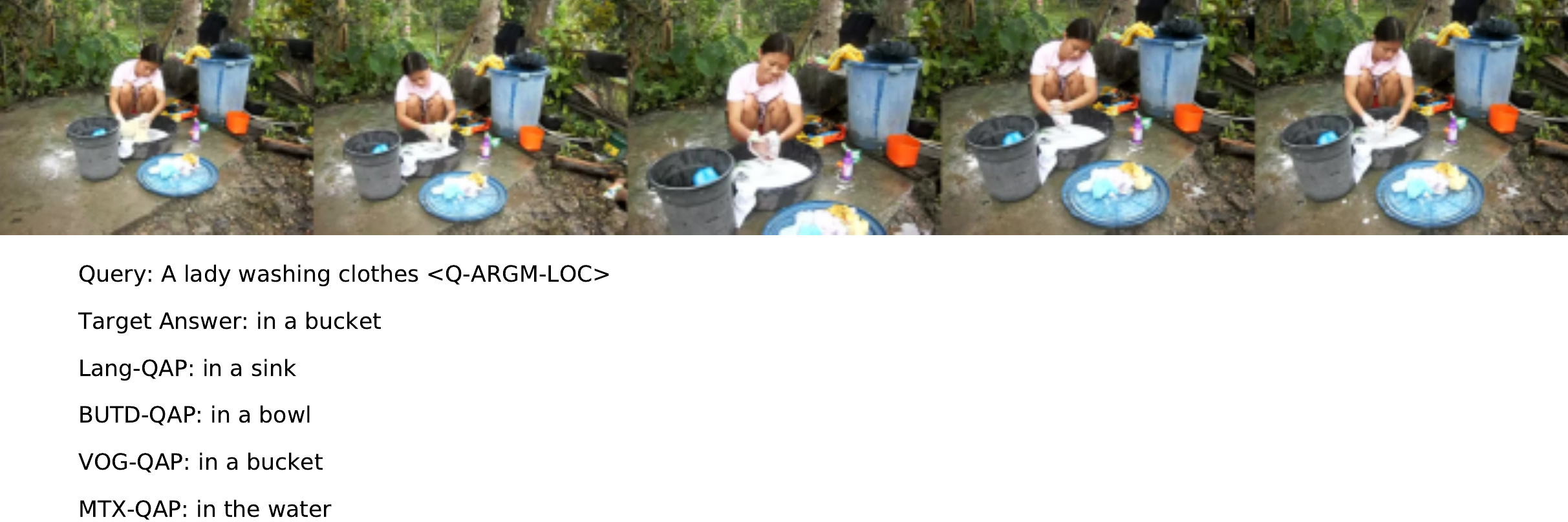}
    \includegraphics[width=\linewidth]{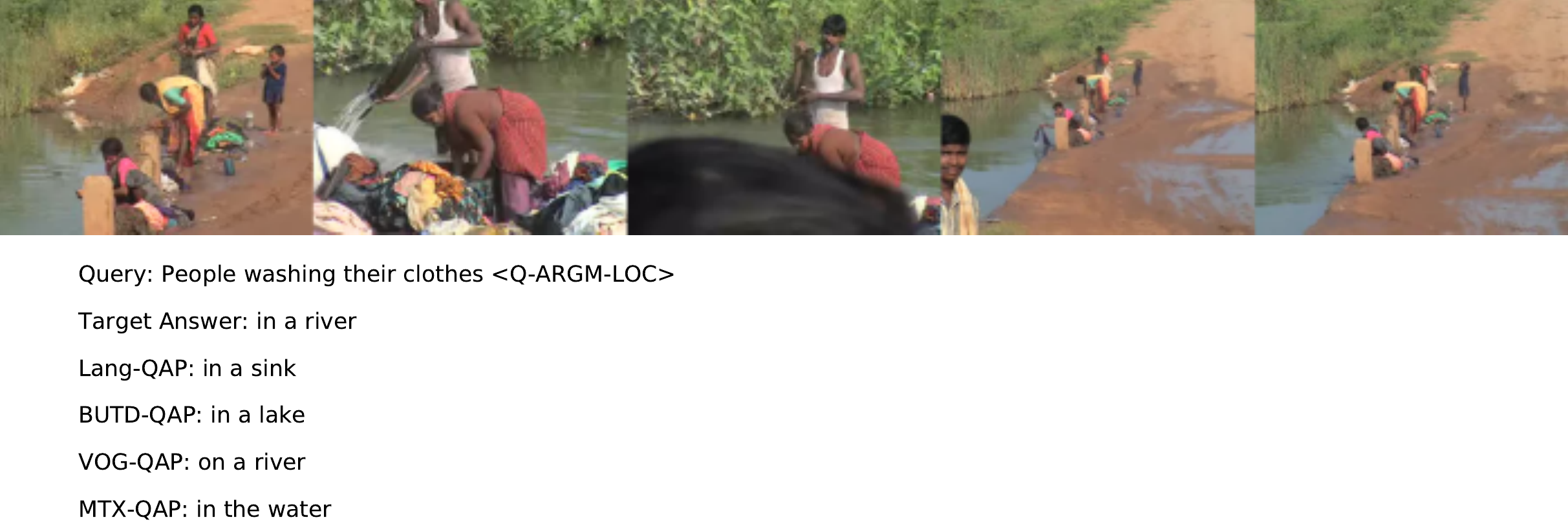}
    \caption{Queries of Type \txi{ARGM-LOC} on \anetdsn{}}
    \label{fig:suppl_aloc_anet_vis}
    
\end{figure*}

We visualize the model outputs on \anetdsn{} in Figure \ref{fig:suppl_a0_v_anet_vis} (a), (b), Figure \ref{fig:suppl_a1_a2_anet_vis} (a), (b) and Figure \ref{fig:suppl_aloc_anet_vis}.
For each case, we show the considered input in the first row, and the contrastive sample in the second row. 
Each row contains $5$ frames uniformly sampled from the video segment to be representative of the content observed by the model.
For every query, we show the ground-truth answer and the outputs from \tnamem{Lang}, \tnamem{BUTD}, \tnamem{VOG} and \tnamem{MTX}.

Overall, we often find \tnamem{Lang} suggesting very probable answers, but as expected they are not grounded in the video. 
As a result, in either of the original sample or the contrastive sample, it performs poorly.

\newpage

\end{document}